\documentclass[twoside]{article}

\usepackage[accepted]{aistats2021}
\usepackage[round]{natbib}

\usepackage[T1]{fontenc}    
\usepackage{hyperref}       
\usepackage{url}            
\usepackage{booktabs}       
\usepackage{amsfonts}       
\usepackage{nicefrac}       
\usepackage{microtype}      

\usepackage{multicol}
\usepackage[linesnumbered,ruled]{algorithm2e}
\usepackage[toc,page]{appendix}

\newcommand{\abs}[1]{ \left\lvert#1\right\rvert} 
\newcommand{\norm}[1]{\left\lVert#1\right\rVert} 
\newcommand\given{\;\middle|\;}

\newcommand\independent{\protect\mathpalette{\protect\independenT}{\perp}}
\def\independenT#1#2{\mathrel{\rlap{$#1#2$}\mkern2mu{#1#2}}}

\newcommand{\subf}[2]{%
  {\tiny\begin{tabular}[t]{@{}c@{}}
  #1\\#2
  \end{tabular}}%
}

\usepackage{anyfontsize}
\usepackage{enumerate}
\usepackage{bm} 
\usepackage{xcolor}
\usepackage{tikz} 

\usepackage{graphicx}

\usepackage{wrapfig}

\usepackage{mathtools}
\usepackage{amsmath, amsfonts, amssymb, amsbsy, amsthm}
\theoremstyle{plain}

\theoremstyle{remark}
\newtheorem{remark}{Remark}[section]

\usepackage[font=scriptsize,labelfont=bf]{caption}
\usepackage{anyfontsize}

\usepackage{enumitem}
\setitemize{noitemsep,topsep=0pt,parsep=0pt,partopsep=0pt}
\usepackage[mathscr]{euscript}

\begin{document}

%
\runningtitle{WRF and Applications in HTE}

%
\runningauthor{Du, Biau, Petit, Porcher}

\twocolumn[

\aistatstitle{Wasserstein Random Forests\\ and Applications in Heterogeneous
Treatment Effects}

\aistatsauthor{ Q. Du \;\; G. Biau \And  F. Petit \;\; R. Porcher }

\aistatsaddress{ Sorbonne Université, CNRS, LPSM \\ Paris, France\And Université de Paris, CRESS, INSERM, INRA\\ Paris, France  } ]

\begin{abstract}
We present new insights into causal inference in the context of
Heterogeneous Treatment Effects by proposing natural variants of Random Forests
to estimate the key conditional distributions. To achieve this, we recast
Breiman’s original splitting criterion in terms of Wasserstein distances between
empirical measures. This reformulation indicates that Random Forests are well
adapted to estimate conditional distributions and provides a natural extension
of the algorithm to multi-variate outputs. Following the philosophy of
Breiman’s construction, we propose some variants of the splitting rule that are
well-suited to the conditional distribution estimation problem.  Some
preliminary theoretical connections are established along with various numerical
experiments, which show how our approach may help to conduct more transparent
causal inference in complex situations.
A \texttt{Python} package is also provided.
\end{abstract}

\section{Introduction}

One of the primary objectives of supervised learning is to provide an estimation of the
conditional expectation $\mathbb{E}\left[Y\given X = x\right]$ for some
underlying 1-dimensional objective $Y$ and a multidimensional covariate $X$ given the dataset $\mathcal{D}_n = \{(X_i,Y_i):1\leq i\leq n\}$.
However, in many real-world applications, it is also
important to extract the additional information encoded in the conditional
distribution $\mathcal{L}\left(Y\given X = x\right)$. This is particularly the
case in the field of Heterogeneous Treatment Effects (HTE) estimation problems,
which represent the main motivation of this work.
\subsection{Motivation}
In HTE problems, the traditional object of
interest is the Conditional Average Treatment Effect (CATE) function,
defined by
\begin{equation}
\label{eq:CATE}
\tau(x) = \mathbb{E}\left[Y(1) - Y(0)\given X = x\right],
\end{equation}
where $Y(1)$ (resp.\@ $Y(0)$) denotes the potential outcome (e.g., \citealp{Rub74,IR15}) of the treatment
(resp.\@ no treatment). 
The propensity score function $e(\cdot)$ is defined by 
\[
  e(x) =
  \mathbb{P}\left(T = 1 \given X=x\right),
\]
which captures the probability of receiving the treatment for each individual.
The data are usually of the form $\bar{\mathcal{D}}_n=\{(X_i,Y_i(T_i),T_i): 1\leq i \leq n\}$, where $T_i$
denotes the treatment assignment indicator. 
Recently, many approaches based on
modern statistical learning techniques have been
investigated to estimate the CATE function (e.g.,
\citealp{KSBY17,AW19,NW17}).
Typically, assuming unconfoundedness, that is
\begin{equation}
\label{eq:unconfoundedness}
\left. \left(Y(0),Y(1)\right) \independent T \given X \right.,
\end{equation}
and that the propensity score function is uniformly bounded away from 0 and 1,
one is able to estimate 
$
\mu_0(x)=\mathbb{E}\left[Y(0)\given X = x\right]
$
and
$
\mu_1(x)=\mathbb{E}\left[Y(1)\given X = x\right],
$
respectively with 
\begin{equation}
\label{eq:data-divide-W}
\left\{(X_i, Y(T_i)): T_i = 0\right\}
\;
\text{and}
\;
\left\{(X_i, Y(T_i)):T_i = 1\right\}.
\end{equation}
The classical approach in the HTE context is to design the causal inference procedure
around the estimation of the CATE function $\tau(\cdot)$ defined in \eqref{eq:CATE} using  $\bar{\mathcal{D}}_n$, and to test whether there is a notable difference between $\tau(x)$ and $0$ for each new coming individual $x$.
It is important to note that this is already a difficult task for certain datasets 
due to the unbalance between treatment and control groups or other practical reasons.
For instance, the X-learner \citep{KSBY17} 
is proposed to deal with the unbalanced design by making efficient use of the structural information 
about the CATE function,
and the R-learner \citep{NW17} is introduced to improve accuracy and robustness of the CATE function estimation
by formulating it into a standard loss-minimization problem.
However, a simple inference based on the CATE function, or other key features
composed by the average treatment effects. (e.g., the sorted Group Average Treatment Effects 
(GATES) proposed by \citealt{Chernozhukov}), may be hazardous in some situations because of the lack of information on the fluctuations, or multimodality, of both conditional laws $\mathcal{L}\left(Y(0)\given X = x\right)$
and $\mathcal{L}\left(Y(1)\given X = x\right)$.
This phenomenon, in practice, can arise when $Y(0)$ and/or $Y(1)$ depend on some additional unconfounding factors that are not collected in the study, which, however, greatly affect the behaviors of the potential outcomes. 
From another point of view, being aware of the existence of such problems 
may also help to fix the flaw of data collection procedure or the lack of subgroup analysis for future study.

Ideally, one is interested in estimating the joint conditional distribution
$
\mathcal{L}\left((Y(0), Y(1))\given X = x\right).
$
Unfortunately, a major difficulty of HTE estimation lies in the fact that it is 
in general impossible to collect $Y_i(0)$ and $Y_i(1)$ at the same time for the
point $X_i$.
Unlike the difference in the linear conditional expectation $\tau(\cdot)$, the dependence between 
$Y(0)$ and $Y(1)$ given $X$ is much
more complex and difficult to track. 
Hence, due to the lack of information of 
the collectable dataset, the estimation of
the conditional covariance between $Y(0)$ and $Y(1)$ is usually unavailable,
let alone the conditional joint distribution.
A possible route to address this shortcoming
is to concentrate on a weaker problem: instead of estimating 
$\mathcal{L}\left((Y(0), Y(1))\given X = x\right)$,
we are interested in the estimation of conditional marginal distributions
$\mathcal{L}\left(Y(0)\given X = x\right)$ and
$\mathcal{L}\left(Y(1)\given X = x\right)$. 
By considering the two subgroups \eqref{eq:data-divide-W} of the dataset $\bar{\mathcal{D}}_n$, the problem thus enters into a more classical supervised learning context, similar as the design of T-learners \citep{KSBY17}, while the
objective is replaced by the estimation of
conditional distributions. 
In some scenarios, even a simple raw visualization of the marginal conditional distributions,
as a complement of CATE function estimation,
may greatly help the decision making process for practitioners.

Another motivation comes from the need to set-up statistically sound decision procedures for multivariate objectives in the context of HTE. For example, a treatment is often related to a cost, which is also collectable and sometimes essential to the final treatment decisions. 
In this context, a simple extension of the CATE function will clearly not be able to capture the dependencies between the treatment effects and the cost.
Thus, a statistical tool that allows conditional distribution estimation with multivariate objective will therefore be useful for more complex inferences involving both treatment effects and costs at the same time. 
In general, the traditional nonparametric methods for conditional distribution inference 
(e.g., \citealp{CDE1,CDE2})
are less effective when it comes to flexibility of implementation,
parallelization, and the ability to handle high-dimensional noisy data. 
Another remark is that Gaussian Process-based methods (e.g., \citealp{GPCDE}) usually require the existence of density w.r.t.\@ Lebesgue measure, which is not always true in the context of HTE. 
So, our goal is to achieve density-free conditional distribution estimation based on available modern machine/statistical learning tools.
\subsection{Random Forests for conditional distribution estimation}
In order to address the issues described in the subsection above, our idea is to propose 
an adaptation of the Random Forests (RF) algorithm \citep{RF}, so that it can be applied to the conditional distribution estimation problems in the HTE context.
RF have proven to be successful in many real-world
applications---the reader is referred to \citet{BS15}
and the references therein for a general introduction.
If we look at the final prediction at each point $x$ provided by the RF algorithm,
it can be regarded as a weighted average of $(Y_i:1\leq i\leq n)$,
where the random weights depend upon the training dataset and the stochastic mechanism of the forests.
Therefore, a very natural idea is to use this weighted empirical measure to
approximate the target conditional distribution. This is also the driving force in the construction of Quantile Regression Forests \citep{QRF,ATW19}
and other approaches that combine kernel density estimations and Random
Forests (e.g., \citealp{RFCDE,hothorn2017transformation}). 

In the present article, instead of studying the quantile or density function of 
the target conditional distribution, we focus directly on the (weighted) empirical 
measures output by the forests and the associated Wasserstein distances.
This also makes further inferences based on Monte-Carlo methods or smoothing more
convenient and straightforward. To make it clearer, let us denote by $\pi(x,dy)$ the probability measure associated with the conditional 
distribution $\mathcal{L}\left(Y \given X = x\right)$. Heuristically speaking, 
if the Wasserstein distance between the Markov kernels 
$\pi(x,dy)$ and $\pi(z,dy)$ is dominated, in some sense,
by the distance between $x$ and $z$, then
the data points that fall into a ``neighborhood'' of $x$ are
expected to be capable of providing
reliable approximation of the conditional measure $\pi(x,dy)$. 
In the RF context, the role of each tree in the ensemble is to build a wisely created 
partition of the domain,
so that the ``neighborhood'' mentioned above can be defined accordingly.
As such, the random weights come from the averaging procedure of multiple trees.

As Breiman's original RF are primarily designed for conditional expectation estimations, 
we first provide 
in Section \ref{sec:method} 
a reformulation that gives new insights into
Breiman's original splitting criterion, 
by exploiting a simple relation between empirical variance and 
Wasserstein distance between empirical measures. This reformulation allows a new
interpretation of the RF algorithm in the context of conditional distribution estimation, which, in turn, can be used to handle multivariate objectives with a computational cost that grows linearly with the dimension of the output. We also investigate in this section several dedicated modifications
of Breiman's splitting rule and present some preliminary theoretical connections between 
their constructions.
With a slight abuse of language, all these RF variants aiming at conditional distribution estimation
are referred to as Wasserstein Random Forests (WRF) in this article.
Finally, we return in Section \ref{sec:application} 
to the HTE problem and illustrate through various numerical experiments how WRF may 
help to design more transparent causal inferences in this context.
\section{Wasserstein Random Forests}
\label{sec:method}
In order to simplify the introduction of WRF, we temporarily 
limit the discussion to the classical supervised learning setting. 
Let $X\in\mathbb{R}^d$ and $Y\in \mathbb{R}^{d'}$ be, respectively, the canonical random variables of covariate and the objective. Our goal is to estimate the conditional measure $\pi(x,dy)$ 
associated with $\mathcal{L}\left(Y\given X = x\right)$ using 
the dataset $\mathcal{D}_n = \{(X_i,Y_i):1\leq i\leq n\}$.
\subsection{Mechanism of Random Forests}
A Random Forest is an ensemble method that aggregates a collection of randomized decision trees. Denote by $M$ the number of trees and, for $1 \leq j \leq M$, let $\Theta_j$ be the canonical random variable that captures randomness of the $j$-th tree.
Each decision tree is trained on a randomly selected dataset $\mathcal{D}_n^*(\Theta_j)$ 
with the same cardinal $a_n \in \{2, \hdots, n\}$, sampled uniformly in $\mathcal{D}_n$ with or without replacement.
More concretely, for each tree, a sequence of axis-aligned splits is made recursively 
by maximizing some fixed splitting rule. At each iteration, $\textbf{mtry}\in \{1, \hdots, d\}$ directions are explored and 
the splits are always performed in the middle of two consecutive data points, 
in order to remove the possible ties. 
The splitting stops when the current cell contains fewer points than a 
threshold $\textbf{nodesize}\in \{2,\dots, a_n\}$, or when all the data points are identical. In this way, a binary hierarchical
partition of 
$\mathbb{R}^d$
is constructed. 
For any $x \in \mathbb{R}^d$, we denote by $A_n(x;\Theta_j,\mathcal{D}_n)$ 
the cell in the $j$-th tree that contains $x$ and by $N_n(x;\Theta_j,\mathcal{D}_n)$ the number of data points in $\mathcal{D}_n^*(\Theta_j)$ 
that fall into $A_n(x;\Theta_j,\mathcal{D}_n)$.

The core of our approach relies on 
the fact that the prediction $\pi_n(x,dy;\Theta_j,\mathcal{D}_n)$ of the conditional
distribution at point $x$
given by the $j$-th tree 
is simply the empirical measure associated with the observations that fall into the same cell 
$A_n(x;\Theta_j,\mathcal{D}_n)$ as $x$, that is
\[
\pi_n(x, dy;\Theta_j,\mathcal{D}_n)
=
\sum_{i\in \mathcal{D}_n^*(\Theta_j)}
\frac{
\mathbf{1}_{\{X_i \in A_n(x;\Theta_j,\mathcal{D}_n)\}}
}{
N_n(x;\Theta_j,\mathcal{D}_n)
}
  \delta_{Y_i}(dy),
\]
where $\delta_{Y_i}(dy)$ is the Dirac measure at $Y_i$. Let $\Theta_{[M]}=(\Theta_1, \hdots, \Theta_M)$. As such, the final estimation $\pi_{M,n}(x,dy;\Theta_{[M]},\mathcal{D}_n)$ provided by the forest is but the average of the $\pi_n(x, dy;\Theta_j,\mathcal{D}_n)$, $1\leq j \leq M$, over the $M$ trees, i.e.,
\begin{equation*}
\pi_{M,n}(x,dy;\Theta_{[M]},\mathcal{D}_n)
=\frac{1}{M}\sum_{j = 1}^M
\pi_{n}(x,dy;\Theta_j,\mathcal{D}_n).
\end{equation*}
Equally,
$
\pi_{M,n}(x,dy;\Theta_{[M]},\mathcal{D}_n)
=\sum_{i = 1}^n \alpha_{i}(x) \delta_{Y_i}(dy),
$
where
$
  \alpha_{i}(x)=  \sum_{j = 1}^M
\frac{
\mathbf{1}_{\{X_i \in A_n(x;\Theta_j,\mathcal{D}_n)\}}
}{
MN_n(x;\Theta_j,\mathcal{D}_n)
}
\mathbf{1}_{\left\{i\in \mathcal{D}_n^*(\Theta_j)\right\}}
$
is the random weight associated with $Y_i$. It is readily checked that
$
\sum_{i = 1}^n \alpha_{i}(x) =  1
$ for any $x\in \mathbb{R}^d$.
Thus, the final prediction $\pi_{M,n}(x,dy;\Theta_{[M]},\mathcal{D}_n)$
is a weighted empirical measure with random weights naturally given by the tree aggregation mechanism. 
Our notation is compatible with \citet{BS15}, where a more detailed introduction to RF is provided. It should be stressed again that we are interested in learning the conditional distribution $\mathcal{L}\left(Y\given X = x\right)$, not in inferring the conditional expectation $\mathbb{E}\left[Y\given X = x\right]$ as in traditional forests. This is of course a more complex task, insofar as the expectation is just a feature, albeit essential, of the distribution.



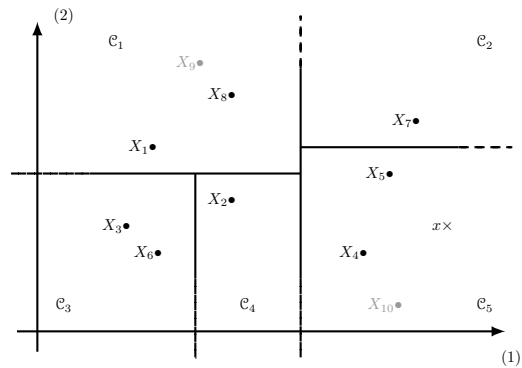
\begin{figure}[htb]
  \centering
\begin{tikzpicture}[thick,scale=0.70, every node/.style={scale=0.60}]

\node at (-2.5,1.5) {$X_1\bullet$};
\node at (-1,0.5) {$X_2\bullet$};
\node at (-3,0) {$X_3\bullet$};
\node at (1.5,-0.5) {$X_4\bullet$};
\node at (2,1) {$X_5\bullet$};
\node at (-2.4,-0.5) {$X_6\bullet$};
\node at (2.5,2) {$X_7\bullet$};
{\color{black!40!white}
\node at (2.1,-1.5) {$X_{10}\bullet$};
}
\node at (-1,2.5) {$X_8\bullet$};
{\color{black!40!white}
\node at (-1.6,3.1) {$X_9\bullet$};
}
\node at (3.2, 0) {$x\times$};
\draw (0.5,3) node (v7) {} rectangle (0.5,-1) node (v8) {};
\node (v1) at (-5,-2) {};
\node (v4) at (-4.5,-2.5) {};
\node (v2) at (4.5,-2) {};
\draw [-latex] (v1) edge (v2);
\node (v3) at (-4.5,4) {};
\draw  [-latex](v4) edge (v3);
\node at (-4,4) {$(2)$};
\node at (4.5,-2.5) {$(1)$};
\draw  (-3.5,1) node (v5) {} rectangle (0.5,1);
\draw  (-1.5,1) rectangle (-1.5,-1) node (v6) {};
\draw  (0.5,1.5) rectangle (3.5,1.5) node (v9) {};
\draw  [dashed](v5) rectangle (-5,1);
\draw  [dashed](v6) rectangle (-1.5,-2.5);
\draw  [dashed](v7) rectangle (0.5,4);
\draw  [dashed](v8) rectangle (0.5,-2.5);
\draw  [dashed](v9) rectangle (4.5,1.5);
\node at (-3,3.5) {$\mathscr{C}_1$};
\node at (4,3.5) {$\mathscr{C}_2$};
\node at (-4,-1.5) {$\mathscr{C}_3$};
\node at (-0.5,-1.5) {$\mathscr{C}_4$};
\node at (4,-1.5) {$\mathscr{C}_5$};
\end{tikzpicture}
\caption{Illustration of the a single decision tree. Note that $X_9$ and $X_{10}$ are not sampled in the sub-dataset used for the tree's construction.}
\label{fig:toy-eg}
\end{figure}
As an illustration, consider in Figure \ref{fig:toy-eg} the partition
$\mathscr{C}_1\cup \mathscr{C}_2 \cup \mathscr{C}_3 \cup \mathscr{C}_4 \cup \mathscr{C}_5 = \mathbb{R}^2$ provided by a decision tree
trained on a bidimensional sub-dataset of size 8. The estimation of the conditional distribution
at the point $x$ is simply
the empirical measure $\frac{1}{2}(\delta_{Y_4}+\delta_{Y_5})$ associated with 
the cell $\mathscr{C}_5$ to which it belongs. Mutatis mutandis, suppose that there is another
decision tree that gives the measure $\frac{1}{2}(\delta_{Y_5}+\delta_{Y_7})$ as the estimation at $x$. Then the final estimation of the conditional distribution output by the forest that contains these two trees is the empirical distribution $\frac{1}{4} \delta_{Y_4} + \frac{1}{2}\delta_{Y_5} + \frac{1}{4}\delta_{Y_7}$.
On the other hand, the classical RF outputs the average $\frac{1}{4} Y_4 + \frac{1}{2}Y_5 + \frac{1}{4}Y_7$ as the scalar estimation of the conditional expectation.
\subsection{Breiman's Splitting criteria} 
\label{sec:splitting}
Now, let us take a closer look at the splitting criteria that are maximized 
at each cell in the construction of trees.
For a cell that consists in a subset of data points $A \subset \mathcal{D}_n$, an axis-aligned cut along the $k$-th coordinate
at position $z$ defines a partition $A_L \cup A_R$ of $A$. More precisely, we
denote
\[
A_L = \left\{
(X_i,Y_i) \in A
:
X_i^{(k)} \leq z
\right\}
\]
and
\[
A_R = \left\{
(X_i,Y_i)\in A
:
X_i^{(k)} > z
\right\}.
\]
With a slight abuse of notation, we write $X_i \in A$
when $(X_i,Y_i) \in A$. 
Recall that Breiman's original splitting criterion \citep{RF} takes the following form:
\begin{equation}
    \label{eq:breiman1}
    \begin{aligned}
      &
L_B(A_L,A_R)
=
\frac{1}{N_A} \sum_{X_i \in A} (Y_i - \bar{Y}_A)^2
\\&-
\frac{1}{N_A} \sum_{X_i\in A_L} (Y_i - \bar{Y}_{A_L})^2
-
\frac{1}{N_A} 
\sum_{X_i\in A_R} (Y_i - \bar{Y}_{A_R})^2,
    \end{aligned}
\end{equation}
where $\bar{Y}_A$ (resp. $\bar{Y}_{A_L}$, $\bar{Y}_{A_R}$) is the average of
the $Y_i$ that fall into $A$ (resp.\@ $A_L$, $A_R$), and $N_A$ (resp.\@ $N_L$,
$N_R$) is the cardinal of $A$ (resp.\@ $A_L$, $A_R$). This criterion is
maximized at each node of each tree over $z$ and the \textbf{mtry} randomly
chosen coordinates (see, e.g., \citealp[section 2.2]{BS15}).
  \begin{remark}
    \label{rmk:breiman1}
  We recall that the Breiman's splitting criterion can be regarded as the
  gain induced by the cut in terms of empirical $L^2$-error. 
 Hence, the 
  mechanism of the construction of the decision tree
  can indeed be interpreted as a greedy optimization.
  \end{remark}
The quantity $L_B$
can also be interpreted as the difference between the total variance and the
intra-class variance within the subgroups divided by the split, which, thanks
to the total variance decomposition, turns out to be the associated inter-class
variance, i.e.,
\begin{equation}
    \label{eq:breiman2}
L_B(A_L,A_R)
=
\frac{N_L}{N_A}(\bar{Y}_{A_L}- \bar{Y}_{A})^2
+
\frac{N_R}{N_A}(\bar{Y}_{A_R}- \bar{Y}_{A})^2.
\end{equation}

  \begin{remark}
    \label{rmk:breiman2}
    As mentioned in Remark \ref{rmk:breiman1}, the representation
    \eqref{eq:breiman2} can be understood as the difference between the predictions with
    and without the cut, in terms of empirical $L^2$-error.  In fact, for each
    data point at $A_L$ (resp. $A_R$), the prediction with the split is
    $\bar{Y}_{A_L}$ (resp. $\bar{Y}_{A_R}$). At the same time, for all the data
    points in $A$, the predictions without the split are given by
    $\bar{Y}_A$. As such, the associated weights $N_L/N_A$ (resp. $N_R/N_A$) come 
    from the number of data points in the sub-cell $A_L$ (resp. $A_R$).
    This reveals a different understanding of the Breiman's splitting rule as
    mentioned in Remark \ref{rmk:breiman1}:
    When considering conditional expectation estimation, the split should be
    made such that the predictions given by the tree are as different as
    possible, in terms of empirical $L^2$-error.  
  \end{remark}
Regardless of the choice of interpretation, 
since there is only a finite number of cuts to be evaluated at each iteration,
a decision tree can therefore be built in a greedy manner. 
Without loss of generality,
when bootstrap is involved (i.e., $\mathcal{D}_n^*(\Theta_j)$ is sampled with replacement), 
one may consider multisets/bags in order to deal with duplicate data for formal 
definitions discussed above.
The details can be
found in Algorithm \ref{algo:wrf}.

\subsection{Basic properties of Wasserstein distances}
Before proceeding further, we recall some basic properties of Wasserstein distances.
If not mentioned otherwise, $d$ and $d'$ denote respectively the dimension of the
covariate $X$ and the dimension of the objective $Y$.
For $p\geq 1$, the Wasserstein distance $\mathscr{W}_p$ between two probability measures $\mu$, $\nu$ on $\mathbb{R}^{d'}$
is defined by
\begin{equation}
  \label{eq:Wp-def}
\mathscr{W}_p(\mu,\nu)
=
\left(
\inf_{
\gamma\in \Gamma(\mu,\nu)
}
\int
\norm{x-y}^p
\gamma(dx,dy)
\right)^{\frac{1}{p}},
\end{equation}
where $\norm{\cdot}$ is the Euclidean norm on $\mathbb{R}^{d'}$ and 
$\Gamma(\mu,\nu)$ denotes the set of all couplings of $\mu$ and $\nu$,
namely,
$
\gamma(dx,\mathbb{R}^{d'}) = \mu(dx)
$ and
$
\gamma(\mathbb{R}^{d'},dy) = \nu(dy)
$.
To guarantee that \eqref{eq:Wp-def} is well-defined,
it is necessary to assume that the $p$-th moments of both $\mu$ and $\nu$ are finite.
When $\mu$ and $\nu$ are the probability measures on $\mathbb{R}$ (i.e., $d'=1$), one can deduce that (see, e.g., \citealp{santambrogio2015optimal})

\begin{equation}
  \label{eq:1d-Wp}
\mathscr{W}_p(\mu,\nu)
=
\left(
\int_0^1
\abs{F_{\mu}^{-1}(u) - F_{\nu}^{-1}(u)
}^p
du
\right)^{\frac{1}{p}},
\end{equation}
where $F^{-1}_{\mu}(u)$ (resp.\@ $F^{-1}_{\nu}(u)$) is the generalized inverse
distribution function defined by
\[
F^{-1}_{\mu}(u) = \inf\left\{
  x\in \mathbb{R} \given F_{\mu}(x) \geq u
\right\},
\]
with $F_{\mu}(x)$ (resp.\@ $F_{\nu}(x)$) the cumulative distribution function of $\mu$ (resp.\@ $\nu$).
Thanks to \eqref{eq:1d-Wp}, the Wasserstein
distance between empirical measures can be efficiently computed in the univariate case.
%
\subsection{Intra-class interpretation}
\label{sec:intra}
We focus on the representation of $L_B$ given in \eqref{eq:breiman1}.
Denote by $\mu_N = \frac{1}{N}\sum_{i=1}^N \delta_{U_i}$.
Observe that
\begin{equation*}
  \mathscr{W}_p^p(\mu_N,\delta_{V_1})
=
\frac{1}{N}
\sum_{i=1}^{N}
\norm{U_{ i}-V_{1}}^p
,
\end{equation*}
one can rewrite the Breiman's rule 
in terms of quadradic Wasserstein distances between empirical measures, namely,
\begin{equation}
  \label{eq:BW}
  \begin{aligned}
    &
L_B(A_L,A_R) 
=\frac{1}{2N_A} 
\sum_{X_i \in A}
\mathscr{W}_2^2\left(\delta_{Y_i},\pi_A\right) 
  \\&
-
\frac{1}{2N_A} 
\sum_{X_i\in A_L} 
\mathscr{W}_2^2\left(\delta_{Y_i},\pi_L\right)
-
\frac{1}{2N_A}\sum_{X_i\in A_R} 
\mathscr{W}_2^2\left(\delta_{Y_i},\pi_R\right) 
.
  \end{aligned}
\end{equation}
In the same spirit of Remark
\ref{rmk:breiman1}, the interpretation \eqref{eq:BW} heuristically indicates that RF are well-adapted to estimate conditional distributions.
More precisely, the Breiman's rule can also be regarded as the gain in terms of
quadratic Wasserstein error, induced by the split.

An important consequence of this result is that it
allows 
a natural generalization of Breiman's criterion 
to outputs $Y$ with a dimension greater than 1. 
Indeed, the extension of RF to multivariate 
outputs is not straightforward, even in the context of conditional expectation estimation (e.g., \citealp{MRF1,MRF2}). The dependence between the different coordinates of the objective is usually dealt with using additional
tuning or supplementary prior knowledge. Such a modeling is not necessary in our approach since dependencies in the $Y$-vector features are captured by the Wasserstein distances. (Note however that some appropriate normalization 
should be considered when there are noticeable differences between the coordinates of the objective.) 
Besides, this extension is also computationally efficient, as the complexity of the evaluation
at each cell increases linearly w.r.t.\@ the dimension $d'$ of the objective
$Y$. The details are provided in Algorithm \ref{algo:L_intra}.
\begin{algorithm}[htb]
\SetAlgoLined
\SetKwInput{KwRequire}{Require}
\KwRequire{Sub-datasets $A_L$, $A_R$ and $A$. }
\KwResult{The value of $L_{\mathrm{intra}}^2(A_L,A_R)$.}

  \For{$k\in\{1,2,\dots, d'\}$}
  {
    Compute respectively 
    $\bar{Y}_{L}^{(k)} = \frac{1}{N_L}\sum_{X_i\in A_L}
    Y_i^{(k)}$, 
    $\bar{Y}_{R}^{(k)} = \frac{1}{N_R}\sum_{X_i\in A_R}
    Y_i^{(k)}$ and
    $\bar{Y}_{A}^{(k)} = \frac{1}{N_A}\sum_{X_i\in A}
    Y_i^{(k)}$. 

    Set respectively $W_{L}^{(k)} = \frac{1}{N_A}\sum_{X_i\in A_L} (Y_{i}^{(k)} -
    \bar{Y}_L^{(k)})$, 
$W_{R}^{(k)} = \frac{1}{N_A}\sum_{X_i\in A_R} (Y_{i}^{(k)} - \bar{Y}_R^{(k)})$
and
$W_{A}^{(k)} = \frac{1}{N_A}\sum_{X_i\in A} (Y_{i}^{(k)} - \bar{Y}_A^{(k)})$.
 }
 Compute and output
 $
 \sum_{k=1}^{d'} \left(W_{A}^{(k)} - W_{L}^{(k)} - W_{R}^{(k)}\right)
 $.

 \caption{Computation of $L_{\mathrm{intra}}^2(A_L,A_R)$ in the case $Y =
   (Y^{(1)},Y^{(2)}, \dots, Y^{(d')})\in
 \mathbb{R}^{d'}$.}
 \label{algo:L_intra}
\end{algorithm}
In the sequel, to increase the clarity of our presentation, we use the notation
$L_{\mathrm{intra}}^2$ instead of $L_B$ for the criterion defined in
\eqref{eq:BW}.

\subsection{Inter-class interpretation}
Using the similar idea as mentioned in Remark \ref{rmk:breiman2} by replacing the $L^2$-error 
with the $\mathscr{W}_2$-distance between empirical measures according to the goal of conditional distribution estimation,
it is natural to consider the following splitting criterion:
\[
\label{eq:inter}
  L_{\mathrm{inter}}^{p}(A_L,A_R)
=
\frac{N_L}{N_A}
\mathscr{W}_p^p(\pi_L,\pi_A)
+
\frac{N_R}{N_A}
\mathscr{W}_p^p(\pi_R,\pi_A).
\]
\begin{remark}
\label{rmk:inter}
A very noticeable difference between intra-class and inter-class interpretation is that
one does not need to choose a reference conditional distribution in the latter case. 
More precisely, at each data point $(X_i,Y_i)$, we have used the reference
conditional distribution $\delta_{Y_i}$ to build the associated local optimizer
in terms of quadratic Wasserstein distance in \eqref{eq:BW}.
Such choice may not be informative enough in the case where the conditional distribution is, say, multimodal. 
However, in the inter-class interpretation, there is no such problem.
\end{remark}
In the univariate case (i.e., $d'=1$), thanks to \eqref{eq:1d-Wp},
it is easily checked that $L_{\mathrm{inter}}^p$
can be computed with $\mathscr{O}(N_A \log(N_A))$ complexity 
at each cell that contains the data points $A\subset \mathcal{D}_n$.
This rate can be achieved by considering a Quicksort algorithm in order to deal with the 
generalized inverse distribution function encountered in \eqref{eq:1d-Wp}. 
The implementation is tractable, although slightly worse than 
$\mathscr{O}(N_A)$, the complexity of $L_{\mathrm{intra}}^2$.
However, the computation of the Wasserstein distance is not trivial when 
$d'>1$, where an exact computation is of 
order $\mathscr{O}(N_A^3)$ (e.g., \citealp[Section 2]{Wp-book}). 
A possible relaxation is 
to consider an entropic regularized approximation such as 
Sinkhorn distance (e.g., \citealp{sinkhorn_nips,sinkhorn_sample,greenkhorn}), where the associated 
complexity is of order $\mathscr{O}(N_A^2\slash \epsilon^2)$ with  tolerance $\epsilon\in (0,1)$. 
Nevertheless, since the amount of evaluations of $L_{\mathrm{inter}}^p$ is enormous during the construction of RF, 
we only recommend using this variant of splitting criterion for univariate problems at the moment.  
The details of efficient implementations and possible relaxations for multivariate cases will be left for future research.
It is however now time to put our splitting analysis to good use and return to the HTE conditional distribution estimation problem.
\begin{algorithm*}[htb]
\SetAlgoLined
\SetKwInput{KwRequire}{Require}
\KwRequire{Training dataset $\mathcal{D}_n$, number of trees $M>0$, subsample
size $a_n \in [n]$, Wasserstein order $p>0$, 
$\textbf{mtry} \in [d]$ where $d$ denotes the dimension of
the covariate $X$, $\textbf{nodesize} \in [a_n]$ and $x\in \mathbb{R}^d$.}
\KwResult{The sequence of weights $\left(\alpha_i(x); i\in [n]\right)$ which determines a
weighted empirical measure that estimates the conditional distribution at $x$.}

  \For{$j\in\{1,2,\dots, M\}$}
  {
    Select $a_n$ points uniformly in
    $\mathcal{D}_n$, with or without replacement, as the sub-dataset  $\mathcal{D}_n^*(\Theta_j)$.

    Initiate a binary tree $\mathcal{T}(\Theta_j,\mathcal{D}_n)$ that
    only contains the root $\mathcal{D}_n^*(\Theta_j)$.

    Set $\mathcal{P} = \left(\mathcal{D}_n^*(\Theta_j)\right)$ the ordered list that
    contains the root of the tree.


    \While{$\mathcal{P} \neq \varnothing$}{

    Let $A$ be the first element of $\mathcal{P}$.

    \eIf{$A$ contains less data points than \textbf{nodesize} or if all $X_i
    \in A$ are identical}{

    Remove the cell $A$ from the list $\mathcal{P}$.

  }{
  Select uniformly without replacement, a subset $\mathcal{M}_{try}\subset [d]$
  of cardinality \textbf{mtry}. 

  Select the best split position $z^*$ and the direction $\ell^*$ based on the
  sub-dataset
  $A$ along the coordinates in $\mathcal{M}_{try}$
  that maximizes the selected splitting rule (i.e., $L_{\mathrm{intra}}^2$ or
  $L_{\mathrm{inter}}^p$). 
  Cut $A$ according to the best split. Denote respectively by $A_L$ and $A_R$ the corresponding
  cells.

  Let the left and right children of $A$ be respectively $A_L$ and
  $A_R$, and associated the node $A$ with the split position and direction $(z^*,\ell^*)$.

  Remove the cell $A$ from the list $\mathcal{P}$.

  Concatenate $\mathcal{P}$, $A_L$ and $A_R$.

}
    }
  Compute  
    $
    \alpha_{i,j}(x) := 
    \frac{
    \textbf{1}_{\{X_i \in A_n(x;\Theta_j,\mathcal{D}_n)\}}
    }{
    MN_n(x;\Theta_j,\mathcal{D}_n) 
    }
    \textbf{1}_{N_n(x;\Theta_j,\mathcal{D}_n)>0}
    $
    for each $(X_i,Y_i)\in \mathcal{D}_n^*(\Theta_j)$
     according to
     $\mathcal{T}(\Theta_j,\mathcal{D}_n)$. 
 }
 Compute $\alpha_i(x) = \frac{1}{M}\sum_{j = 1}^M\alpha_{i,j}(x)$
 for each $i\in [n]$.
 \caption{Wasserstein Random Forests predicted distribution at $x\in
 \mathbb{R}^d$.}
 \label{algo:wrf}
\end{algorithm*}

\section{Applications}
\label{sec:application}
Our primary interest is the improvement that WRF can bring into the causal inference under
the potential outcomes framework. 
As for now, the potential outcomes $Y(0)$ and $Y(1)$ are assumed to be univariate random variables---extension to the multivariate case will be discussed a little later. During the observational study, 
the i.i.d.\@ dataset $\bar{\mathcal{D}}_n= \{(X_i,Y_i,T_i):1\leq i \leq n\}$ is collected with $Y_i$ an abbreviation of $Y_i(T_i)$.
Under the
unconfoundedness assumption \eqref{eq:unconfoundedness}, our goal is to
estimate the probability distribution $\pi_t(x,dy)$ associated with the conditional marginal distribution
$\mathcal{L}\left(Y(t)\given X = x\right)$ for $t\in \{0,1\}$, based on the dataset
$\bar{\mathcal{D}}_n$. 

\subsection{Wasserstein Random Forests for HTE}
Before discussing the applications of these conditional marginal distribution estimations,
we would like to stress again that we have no intention to ``replace''
the Average Treatment Effect-based causal inference strategy. On the contrary, our primary motivation is to provide a complementary 
tool so that a more transparent inference can be conducted, by maximizing the usage of available data.
More precisely, we train WRF 
respectively on the treatment and control groups \eqref{eq:data-divide-W}, to estimate respectively
the conditional measures $\pi_0$ and $\pi_1$. These estimations are denoted by 
$\hat{\pi}_0$ and $\hat{\pi}_1$.

First, when potential outcomes are assumed to be univariate, a raw visualization of $\hat{\pi}_0$ and $\hat{\pi}_1$ is always accessible and informative. 
In this way, causality can therefore be visualized by the change 
of the shape of the marginal distributions. 
Next, following a philosophy similar to the CATE function, we propose to assess the changes in the conditional distribution in terms of Wasserstein distance using the criterion
\[
\Lambda_p(x) = \mathscr{W}_p\left(\pi_0(x,dy), \pi_1(x,dy)\right). 
\]
Intuitively speaking, $\Lambda_p(\cdot)$ is capable of capturing certain causal effects 
that are less noticeable in terms of $\tau(\cdot)$. An estimation $\hat{\Lambda}_p(\cdot)$ 
can be obtained as a 
by-product of the estimation of $\hat{\pi}_0$ and $\hat{\pi}_1$. 
For practical implementation, a histogram of the estimation of $\Lambda_p(x)$ can be constructed by
out-of-bag strategy.
Finally, regarding the multivariate output case, we would like to mention that when the cost of the treatment, say $C(1)$, is also
collected in the dataset, WRF can then be used---as we have seen in Subsection \ref{sec:splitting}, without further effort---to provide an estimation of the joint multivariate distribution 
$\mathcal{L}\left((Y(1), C(1))\given X =x\right)$ in order to conduct more complex inferences involving
the costs and the treatment effects at the same time. The same idea also applies to the case where the treatment effects themselves are also multivariate.
\subsection{Univariate conditional density estimation}
\label{sec:univariate}
Since the conditional distribution is in general inaccessible from the real-world datasets, 
we present here a simulation study based on synthetic data to illustrate the performance of WRF in the context of HTE, focusing on the conditional marginal distribution estimation. 
We consider the following model, where $X=(X^{(1)}, \hdots, X^{(d)})$ and the symbol $\mathcal{N}$ stands for the Gaussian distribution:
\begin{itemize}
\item
$X \sim \mathrm{Unif}\left([0,1]^d\right)$ with $d = 50$;
\item
$Y(0) \sim \mathcal{N}(m_0(X),\sigma_0^2(X))$;
\item
  $Y(1)\sim 
\frac{1}{2}\delta_{-1} + \frac{1}{2}\mathcal{N}\left(m_1(X),\sigma_1^2(X)\right)
$; 
\item
$T\sim \mathrm{Bernoulli}\left(\frac{1}{2}\right)$;
\end{itemize}
with
\begin{itemize}
  \item
$
  m_0(x) = 10 x^{(2)}x^{(4)} + x^{(3)} + \exp\left\{ x^{(4)} -2 x^{(1)}\right\}
$;
\item
$
\sigma_0^2(x) = \left\{-x^{(1)}x^{(2)} + 4\left(x^{(3)}\right)^2 \right\}\vee\frac{1}{5} 
$;
\item
$
m_1(x) = 2m_0(x) + 1 - 5 x^{(2)} x^{(5)} 
$;
\item
$
\sigma_1^2(x) = 3 x^{(2)} + x^{(3)} x^{(4)} + x^{(6)}
$.
\end{itemize}
To summarize, the conditional measure $\pi_0(x,dy)$ is unimodal, while
$\pi_1(x,dy)$ is bimodal, composed by a Gaussian and a Dirac at $-1$, and thus the conditional distribution of $Y(1)$ does not have a density w.r.t.\@ Lebesgue measure.
The mixture parameters $(1/2,1/2)$ in $\pi_1$ can be interpreted as an unconfounding factor that is not collected
in the study. The four functions $m_0(\cdot)$, $\sigma_0^2(\cdot)$, $m_1(\cdot)$, and $\sigma_1^2(\cdot)$
have been designed to implement complex dependence between the covariate and the potential outcomes. We note however that the CATE function takes the simple form $\tau(x)=-2.5x^{(2)}x^{(5)}$ and is therefore equal to zero if and only if $x^{(2)}x^{(5)}=0$.
The treatment and control groups are balanced 
due to the symmetrical form of $T$.
This simple setting is usually referred to as
\emph{randomized study}. This choice allows us to avoid the complexity when
dealing with the propensity score function, and we can thus focus on the
conditional distribution estimation. A brief discussion on the influence of
propensity score function can be found in \textbf{Supplementary Material}.

\begin{figure}[htb]
\centering
\begin{tabular}{cc}
\subf{\includegraphics[width=0.2\textwidth]{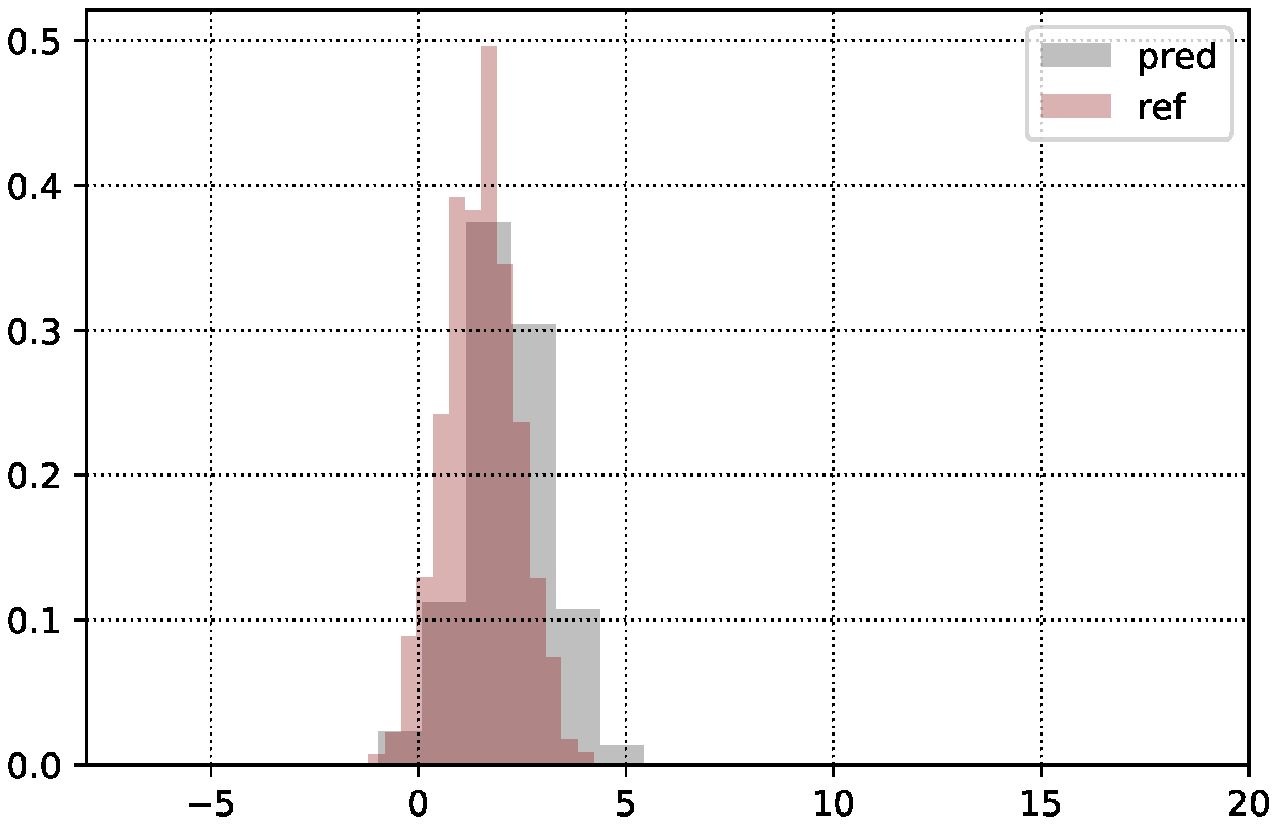}}
{(a) $\pi_0(x_{\ast},\cdot)$ by $L_{\mathrm{intra}}^2$-WRF}
    & 
\subf{\includegraphics[width=0.2\textwidth]{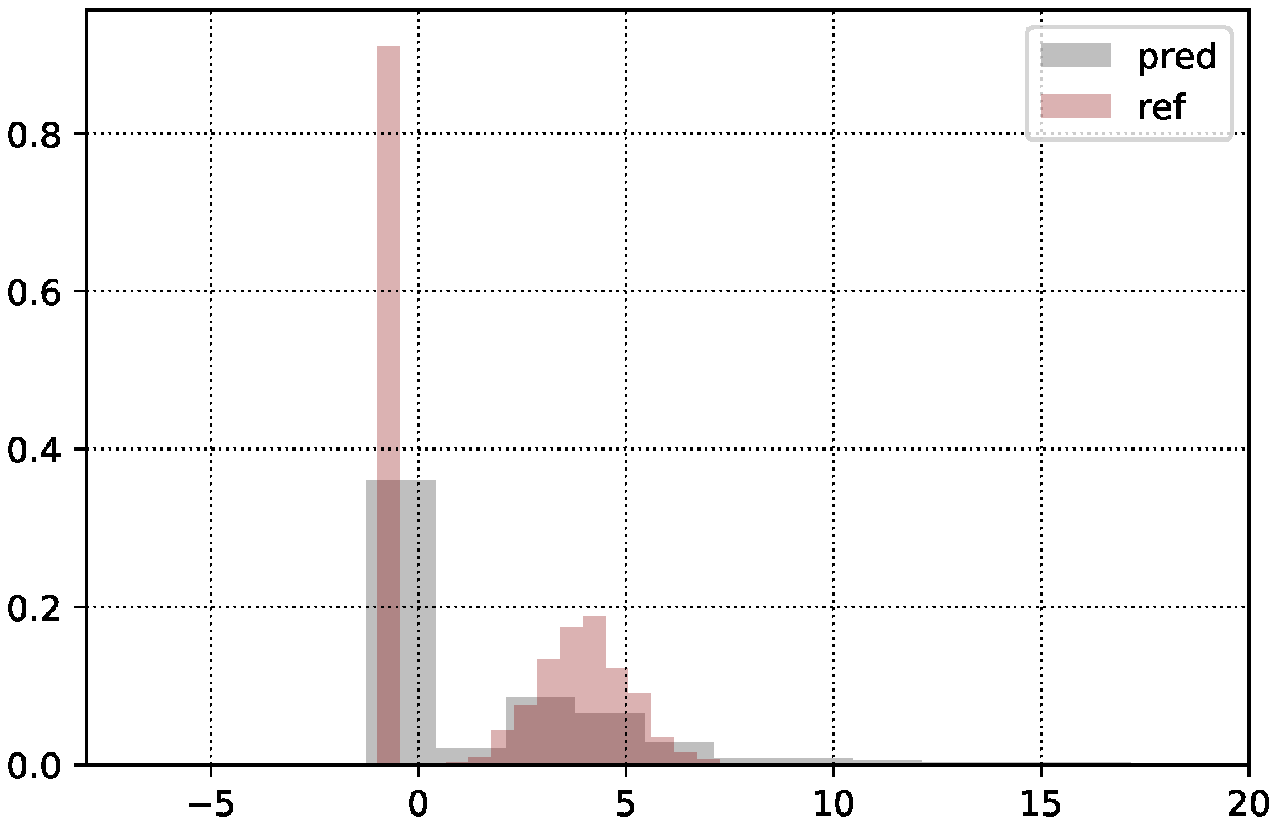}}
     {(b) $\pi_1(x_{\ast},\cdot)$ by $L_{\mathrm{intra}}^2$-WRF}
\\
\subf{\includegraphics[width=0.2\textwidth]{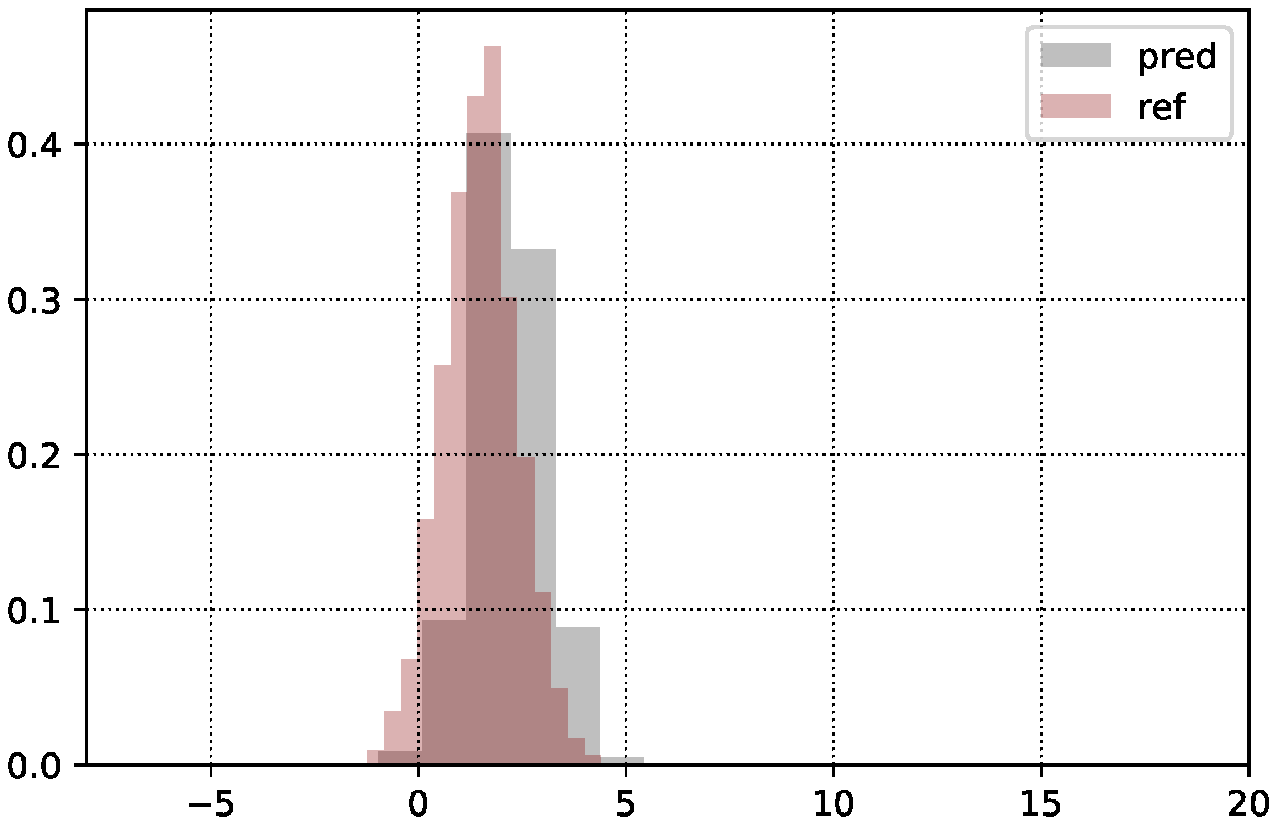}}
     {(c) $\pi_0(x_{\ast},\cdot)$ by $L_{\mathrm{inter}}^2$-WRF}
  &
\subf{\includegraphics[width=0.2\textwidth]{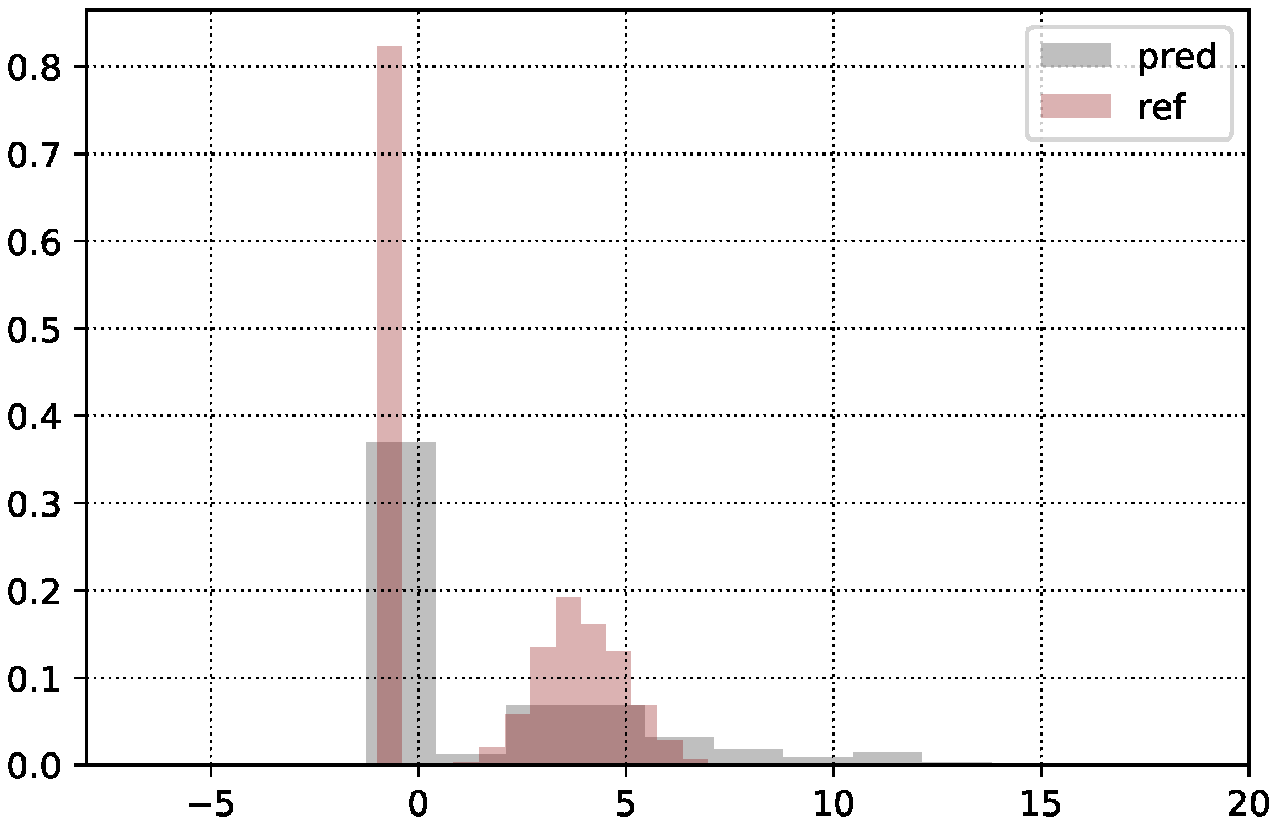}}
     {(d) $\pi_1(x_{\ast},\cdot)$ by $L_{\mathrm{inter}}^2$-WRF}
\\
\end{tabular}
\caption{An illustration of estimated conditional distributions provided 
by different variants of WRF with the same parameters: 
$a_n = 500$ (\text{with repetition}), $M=200$, $\textbf{mtry} = 50$, $\textbf{nodesize}=2$.
In the legend, \texttt{pred} and \texttt{ref} denote respectively the histograms provided by WRF and reference values sampled directly from the true conditional distribution with sample sizes fixed to be 2000.
}
\label{fig:exp}
\end{figure}

We trained the models based on a simulated dataset of size $n=1000$, which is reasonably small
considering the complexity of the conditional distribution estimation problem.
An illustration for an individual $x_{\ast}$ with $x_{\ast}^{(5)} = 0$ (so,
$\tau(x_{\ast})=0$) can be found in Figure \ref{fig:exp} ((a)-(b) for
$L_{\mathrm{intra}}^2$-WRF; (c)-(d) for $L_{\mathrm{inter}}^2$-WRF). This
visualization highlights the good quality of conditional inference performed by
our WRF methods---both of them have highlighted the key properties such as multimodality and fluctuation in the conditional marginal distributions.  
More importantly, it stresses the pertinence of studying
conditional distributions in the HTE context, since the CATE function, as is
the case here, is not always capable to provide insights regarding causality.
For example, according to the trained $L_{\mathrm{intra}}^2$-WRF model, we have
$\hat{\Lambda}_2(x_{\ast}) = 1.8591$ (reference value $\Lambda_2(x_{\ast}) =
2.1903$), which is much more noticeable as an indicator of causality than the
CATE function (estimated by $-0.3675$) in this situation.  

A finer comparison based on average
Wasserstein distance is shown in Table \ref{tab:AWD0} and Table \ref{tab:AWD1}, where
$\pi_t\text{-}\overline{\mathscr{W}}_p(N)$ ($t=0,1$ and $p=1,2$) denotes the
average $\mathscr{W}_p$-distance between $\hat{\pi}_t(x,\cdot)$ and $\pi_t(x,\cdot)$
(approximated by uniformly distributed empirical measures of size 2000) tested on $N$ points (i.e., individuals $x$)
randomly sampled in $[0,1]^d$, which is basically a Monte-Carlo approximation of
$\mathbb{E}\left[\mathscr{W}_p(\hat{\pi}_t(X,\cdot),\pi_t(X,\cdot))\given \bar{\mathcal{D}}_n\right]$.
We compare several WRFs with other
popular RF-based methods that are able to perform conditional distribution estimation.
First, we consider
Mondrian Forests (MF, \citealp{MF}), whose splits do not depend on the response variable.
The idea is to prove the relevance of our splitting criteria in the high-dimensional setting.
Second, we consider Extreme Randomized Trees (ERT, \citealp{ERT}) with Breiman's rule.
This can be seen as $L_{\mathrm{intra}}^2$-WRF with different stochastic construction---the candidates of splitting positions are sampled uniformly on the edges of cells, and then the candidate with
the best score w.r.t.\@ $L_{\mathrm{intra}}^2$ is chosen. 
Finally, we also compare with another RF-based conditional distribution estimation method with different splitting rule (RFCDE, \citealp{RFCDE}) based on $L^2$-error of density estimations.
We stress again that WRF do not need to assume that the conditional density w.r.t.\@ lebesgue measure exists when dealing with conditional distribution estimation.
Finally, we also compare with the classical Kernel Density Estimation (KDE,
\citealp{rosenblatt1969conditional}) and its Nearest Neighbour counterpart
(NN-KDE, \citealp{biau2015new}).
We consider Gaussian kernel, with the smoothing bandwidth parameter $h$ chosen by cross-validated grid-search (for the associated conditional expectation estimation), and the number of neighbours chosen to be the square root of the number of data points.

 \begin{table}[htb]
  \caption{Estimation of $\pi_0$ (i.e., $\mathcal{L}\left(Y(0)\given X=x\right)$)} 
  \label{tab:AWD0}
  \centering
  \begin{tabular}{lll}
    \toprule
    Methods  & $\pi_0\text{-}\overline{\mathscr{W}}_1(1000)$ 
    &$\pi_0\text{-}\overline{\mathscr{W}}_2(1000)$
    \\
    \midrule
        $L_{\mathrm{intra}}^2$-WRF      &0.7209 &0.8809\\
        $L_{\mathrm{inter}}^2$-WRF      &0.7150 &0.8766 \\ 
        $L_{\mathrm{inter}}^1$-WRF      &\textbf{0.7097} &\textbf{0.8642}\\ 
        MF      &2.0835 &2.4576\\
        ERT      &0.7736 &0.9769\\ 
        RFCDE     &0.8111 &0.9725 \\ 
        KDE       & 2.0110       & 2.5321  \\
        NN-KDE       & 1.9250       & 2.3961 \\
    \bottomrule
  \end{tabular}
\end{table}

 \begin{table}[htb]
  \caption{Estimation of $\pi_1$ (i.e., $\mathcal{L}\left(Y(1)\given X=x\right)$)} 
  \label{tab:AWD1}
  \centering
  \begin{tabular}{lll}
    \toprule
    Methods  
    &$\pi_1\text{-}\overline{\mathscr{W}}_1(1000)$
    &$\pi_1\text{-}\overline{\mathscr{W}}_2(1000)$
    \\
    \midrule
    $L_{\mathrm{intra}}^2$-WRF      &1.7030 &2.8111\\
    $L_{\mathrm{inter}}^2$-WRF       &1.3767 & \textbf{2.2987}\\ 
    $L_{\mathrm{inter}}^1$-WRF      & \textbf{1.3498} &2.3341\\ 
    MF      &2.2553 &3.3778\\
        ERT   &  1.6021 & 2.6742\\ 
        RFCDE     &3.2960 &3.4895 \\ 
        KDE       & 2.3958       & 3.2993 \\
        NN-KDE       & 2.2490       & 3.1223\\
    \bottomrule
  \end{tabular}
\end{table}
According to Table \ref{tab:AWD0} and Table \ref{tab:AWD1}, it is clear that WRFs, especially $L_{\mathrm{inter}}^1$ version, provide promising results for this synthetic dataset. In particular, the good performance of $L_{\mathrm{inter}}^p$-WRF w.r.t.\@ $L_{\mathrm{intra}}^2$-WRF may be connected to the discussion provided in Remark \ref{rmk:inter}.
Since the splits of MF do not depend on $Y_i$, it is easy to understand that most of splits are not
performed at ``good'' directions.
The method RFCDE is in general not easy to tune. Despite the choice of kernel,
we have used grid searching for determining the associated hyper parameters such as \texttt{bandwidth}. 
The poor performance in Table \ref{tab:AWD1} may be explained by the non-existence of a probability density for $\pi_1$.  
It is interesting to note that ERT 
outperforms $L_{\mathrm{intra}}^2$-WRF for the estimation of $\pi_1$, which suggests that there is still room to improve 
the stochastic construction of trees.

\subsection{Multivariate case}
\begin{figure}[htb]
\centering
\begin{tabular}{cc}
\subf{\includegraphics[width=0.2\textwidth]{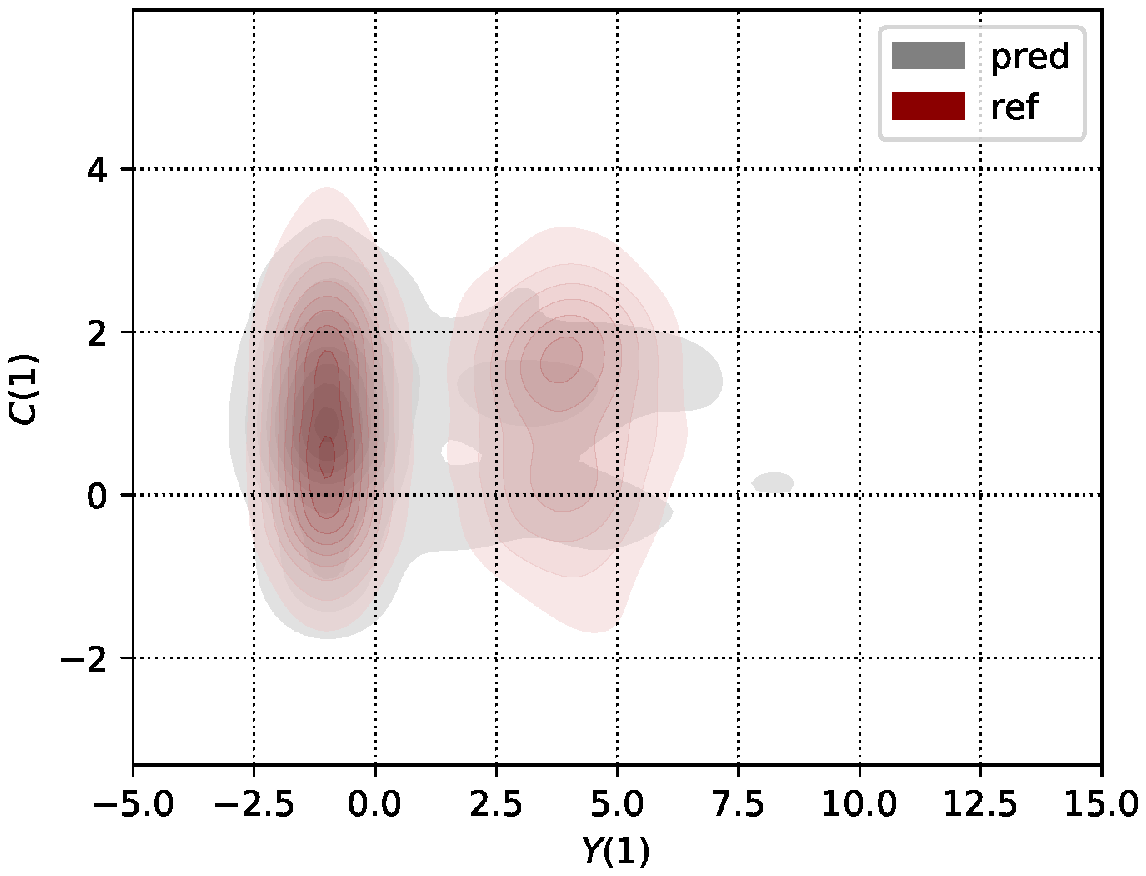}}
{(a) $L_{\mathrm{intra}}^2$-WRF}
    & 
\subf{\includegraphics[width=0.2\textwidth]{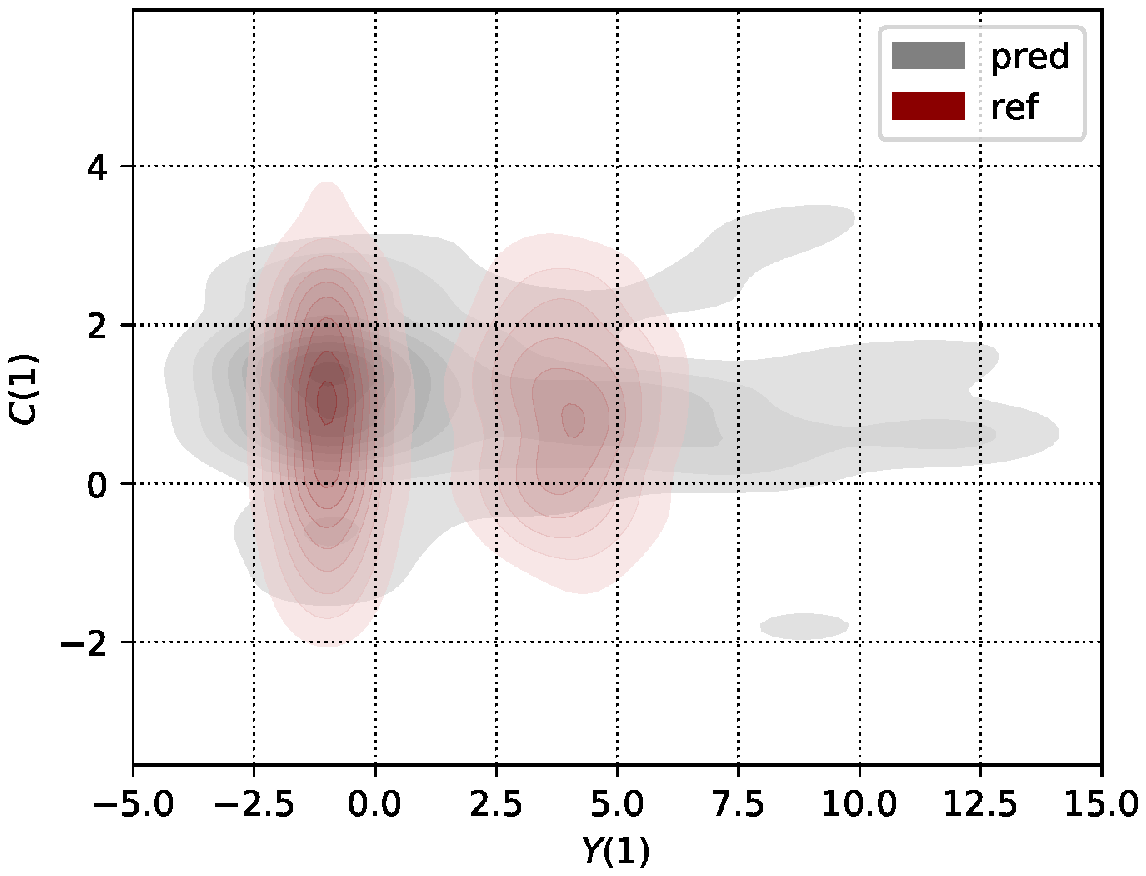}}
     {(b) MF}
\end{tabular}
\caption{
  An illustration of estimated conditional distributions (by heatmap) at a randomly selected
  point provided respectively 
  by $L^2_{\mathrm{intra}}$-WRF and MF.
}
\label{fig:multi}
\end{figure}

We illustrate in this section the ability of $L^2_{\mathrm{intra}}$-WRF
when dealing with multivariate output. 
The implementation can be regarded as a natural generalization of Breiman's rule in the multivariate setting.
Moreover, the complexity is optimal (linear) w.r.t.\@ the dimension of $Y$.
Denote by $C(1)$ the
cost variable associated to the treatment, which is supposed to be a random
variable that depends on $X$, namely,
$
C(1) \sim \mathcal{N}\left(2X^{(3)}X^{(5)}+X^{(2)}, X^{(5)}X^{(6)}+1\right).
$
Our goal is to estimate the joint conditional distribution
$\mathcal{L}\left((Y(1),C(1))\given X=x\right)$. The basic setting of the
algorithm remains the same as discussed in Section \ref{sec:univariate}. 
As shown in Figure \ref{fig:multi}, $L_{\mathrm{intra}}^2$-WRF gives, at least visually, 
a promising estimation of the conditional joint distribution even with only around 500 samples, which outperforms MF (Table \ref{tab:AWD3}). The results, again, provide evidence of the relevance of our Wasserstein distance-based interpretation in the multivariate case. 

 \begin{table}[htb]
  \caption{Estimation of $\mathcal{L}\left((Y(1),C(1))\given X= x\right)$}
  \label{tab:AWD3}
  \centering
  \begin{tabular}{lll}
    \toprule
    Methods  
    &$\pi_1\text{-}\overline{\mathscr{W}}_1(1000)$
    &$\pi_1\text{-}\overline{\mathscr{W}}_2(1000)$
    \\
    \midrule
    $L_{\mathrm{intra}}^2$-WRF      &0.0852 &0.1298\\
        MF      &0.1055 &0.1623\\
    \bottomrule
  \end{tabular}
\end{table}

\section*{Conclusion}
We have proposed a new approach based on WRF that can help HTE inference
through estimating some key conditional distributions.
From a theoretical perspective, the challenge is to prove consistency of WRF when the sample size tends to infinity, in the spirit of works such as \citet{consistency-RF, Wager15}.
For example, a first goal would be to show that, under appropriate assumptions,
$\mathbb{E}\left[\mathscr{W}_p(\hat{\pi}_t(X,\cdot),\pi_t(X,\cdot))\right] \to 0$ as $n \to \infty$,
where $\hat{\pi}_t$ denotes the output of WRF and the expectation is taken
w.r.t.\@ both the distribution of $X$ and the sample.



\section*{Acknowledgements}
We thank the referees for their constructive comments.
This work is funded by the Agence Nationale de la Recherche, under grant agreement no.\@ ANR-18-CE36-0010-01.
Raphaël Porcher acknowledges the support of the French Agence Nationale de la
Recherche as part of the ``Investissements d’avenir'' program, reference
ANR-19-P3IA-0001 (PRAIRIE 3IA Institute).
François Petit acknowledges the support of the Idex “Université de Paris 2019”.

\onecolumn
\begin{appendices}
\section{Additional simulation study for univariate case}
In order to compare with the other conditional density estimation methods such
as RFCDE \citep{RFCDE} and take into account the influence of the propensity score
function, we consider a slightly modified model:
\begin{itemize}
\item
$X \sim \mathrm{Unif}\left([0,1]^d\right)$ with $d = 50$;
\item
$Y(0) \sim \mathcal{N}(m_0(X),\sigma_0^2(X))
\text{ and }
Y(1)\sim 
\frac{1}{2}\mathcal{N}(-1,1) + \frac{1}{2}\mathcal{N}\left(m_1(X),\sigma_1^2(X)\right)
$; 
\item
$T\sim \mathrm{Bernoulli}\left(\frac{1}{2}\sin\left(2X^{(1)}X^{(2)} + 6X^{(3)}\right)+\frac{1}{2}\right)$,
\end{itemize}
with
\begin{itemize}
  \item
$
  m_0(x) = 10 x^{(2)}x^{(4)} + x^{(3)} + \exp\left\{ x^{(4)} -2 x^{(1)}\right\}
$;
\item
$
\sigma_0^2(x) = \left\{-x^{(1)}x^{(2)} + 4\left(x^{(3)}\right)^2 \right\}\vee\frac{1}{5} 
$;
\item
$
m_1(x) = 2m_0(x) + 1 - 5 x^{(2)} x^{(5)} 
$;
\item
$
\sigma_1^2(x) = 3 x^{(2)} + x^{(3)} x^{(4)} + x^{(6)}
$.
\end{itemize}
Basically, the distributions of $X$ and $Y(0)$ remain the same, while the
conditional distribution of $Y(1)$ given $X$ is replaced by a mixture of two
Gaussians, which admits a density w.r.t.\@ Lebesgue measure on $\mathbb{R}^d$.
The propensity score function is also modified in order to model the complexity
of observational studies.

\begin{figure}[h]
\centering
\begin{tabular}{cc}
\subf{\includegraphics[width=50mm]{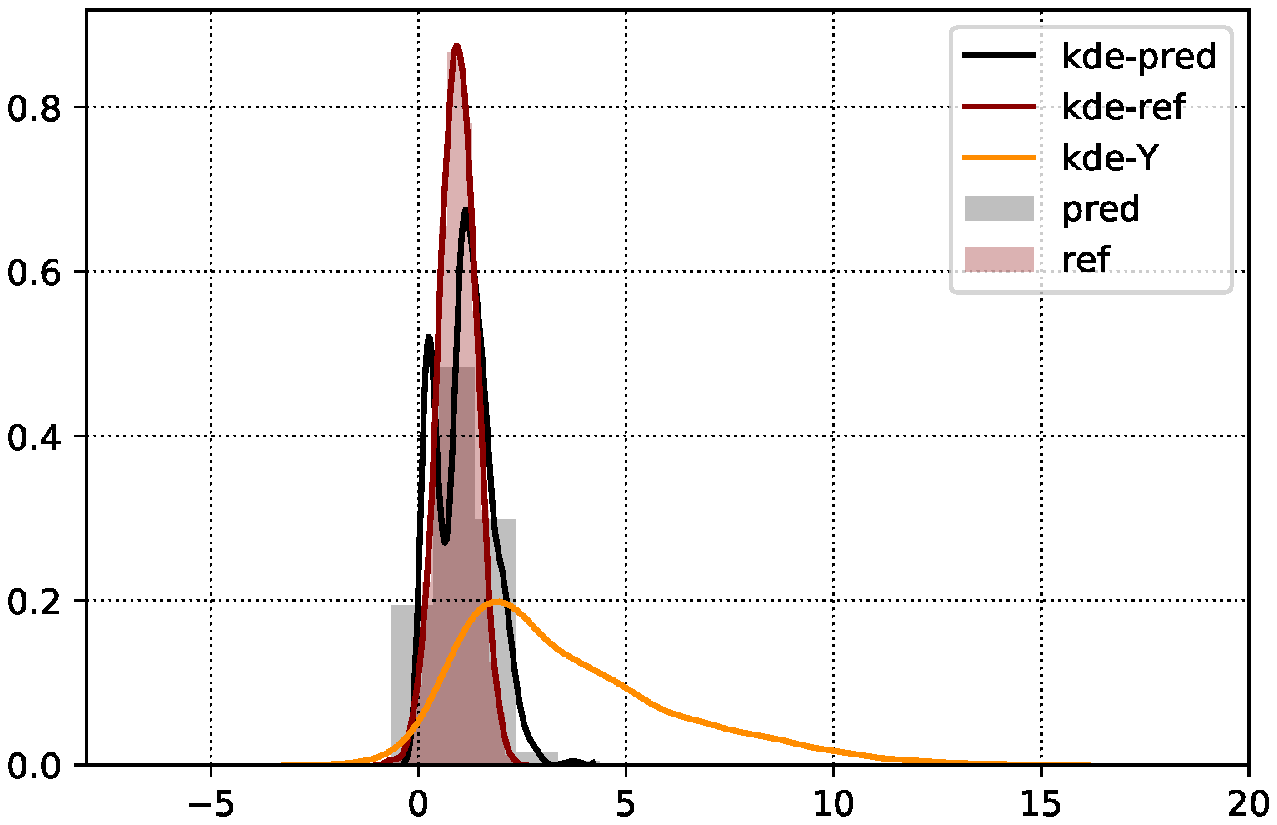}}
     {(a) $\pi_0(x_{\star},\cdot)$ estimated by $L_{\mathrm{intra}}^2$-WRF}
&
\subf{\includegraphics[width=50mm]{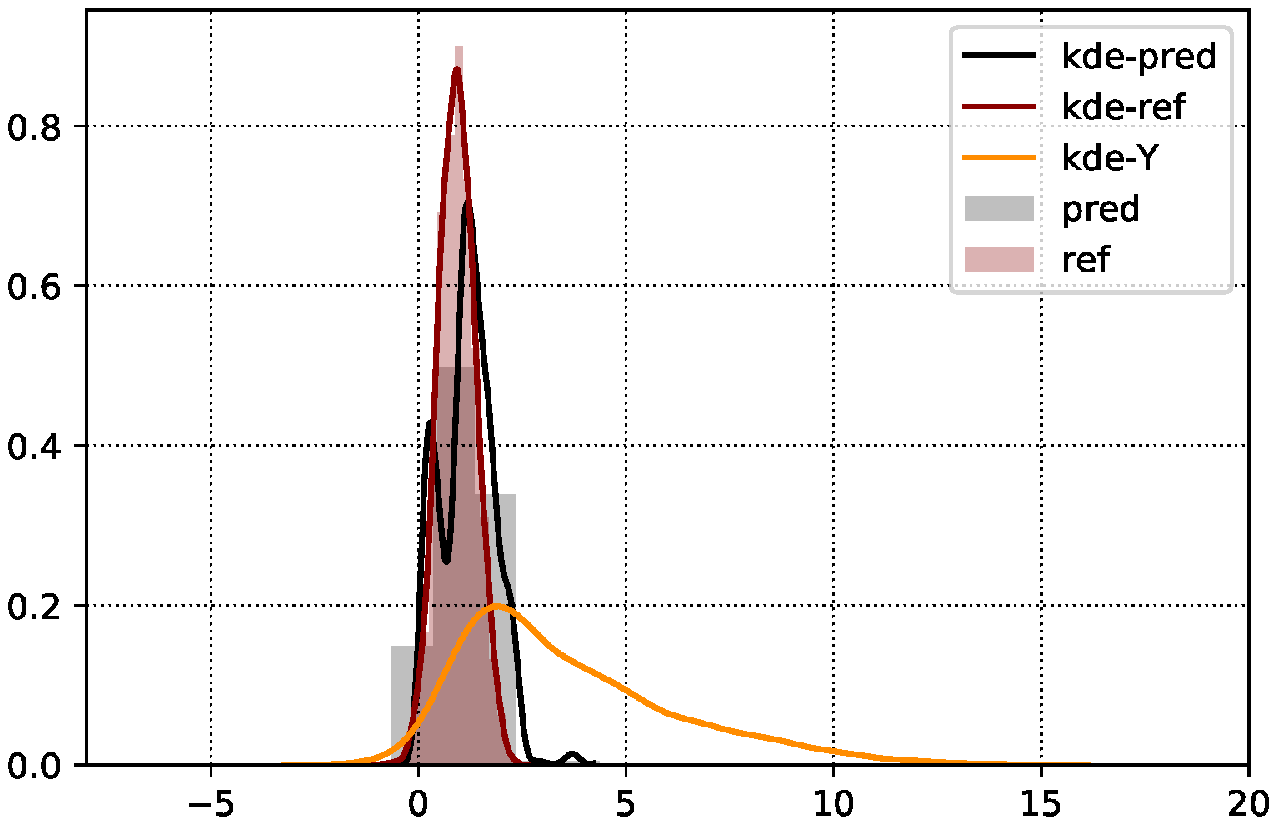}}
     {(b) $\pi_0(x_{\star},\cdot)$ estimated by $L_{\mathrm{inter}}^2$-WRF}
\\
\subf{\includegraphics[width=50mm]{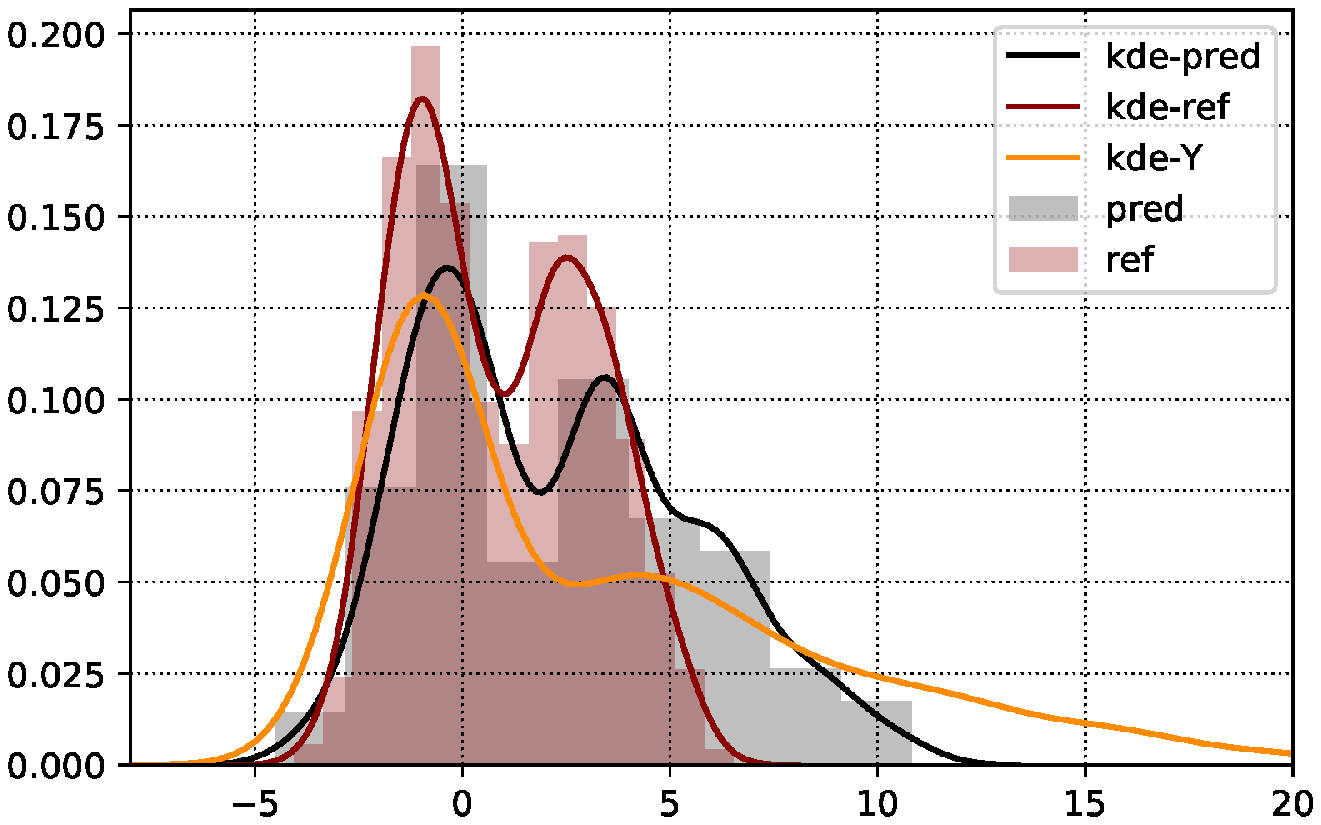}}
     {(c) $\pi_1(x_{\star},\cdot)$ estimated by $L_{\mathrm{intra}}^2$-WRF}
&
\subf{\includegraphics[width=50mm]{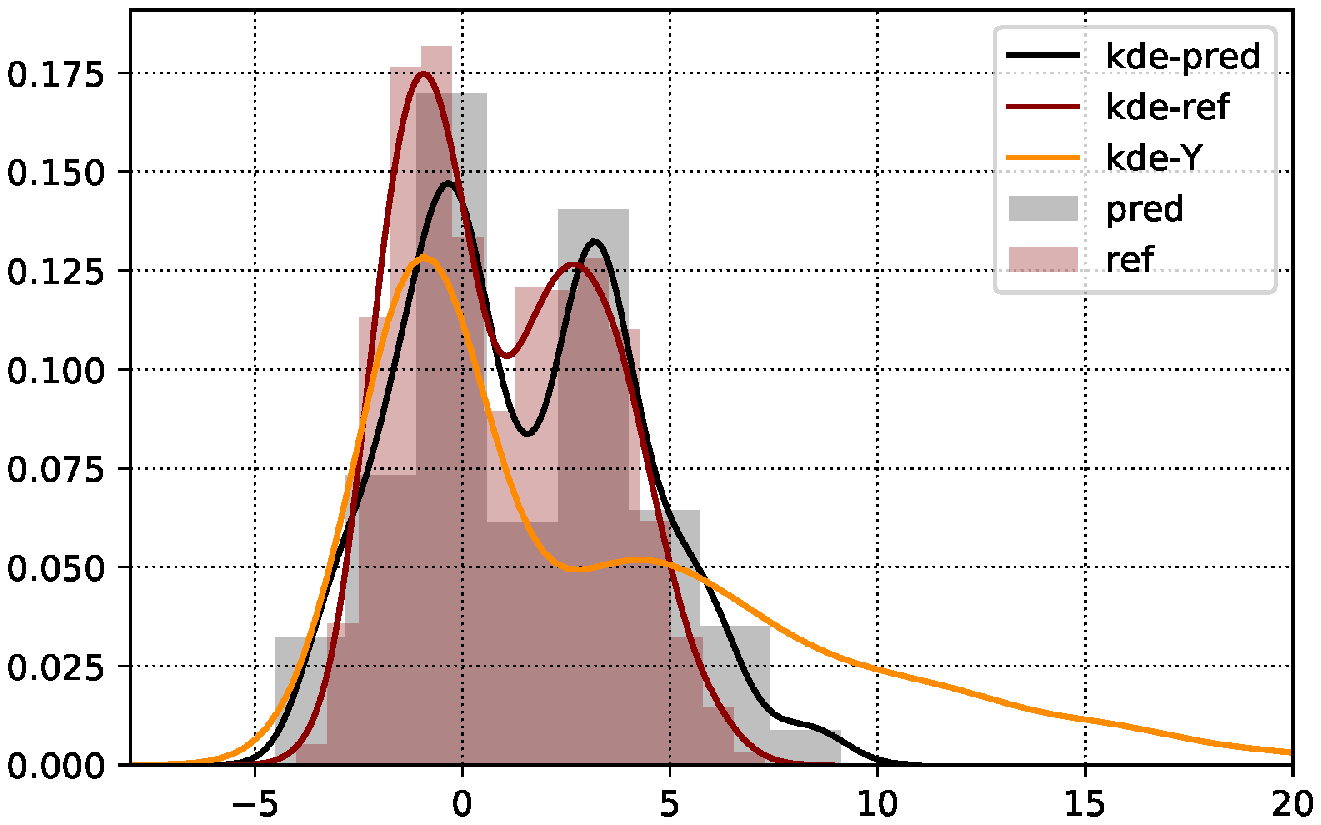}}
     {(d) $\pi_1(x_{\star},\cdot)$ estimated by $L_{\mathrm{inter}}^2$-WRF}
\end{tabular}
\caption{An illustration of estimated conditional distributions provided 
by different variants of WRF with the same parameters: 
$a_n = 500$ (\text{with repetition}), $M=200$, $\textbf{mtry} = 50$, $\textbf{nodesize}=2$.
In the legend, \texttt{pred} and \texttt{ref} denote respectively the prediction given by WRF and reference values sampled directly from the true conditional distribution with sample size fixed to be 2000.
The acronyms \texttt{kde}-\texttt{pred} and \texttt{kde}-\texttt{ref} stand for the outputs of the \texttt{kdeplot} function of \texttt{seaborn} package \citep{seaborn}, which provides a standard kernel smoothing. Finally, \texttt{kde}-\texttt{Y} denotes the \texttt{kdeplot} of the $Y$-population, i.e., all the $Y_i(1)$ or $Y_i(0)$ in the training dataset according to the treatment/control group. 
}
\label{fig:add}
\end{figure}

First, to illustrate the good quality of the estimation provided by WRF, we randomly select an individual $x_{\ast}$ such that
the associated CATE function is 0 (i.e., $x_{\ast}^{(2)}x_{\ast}^{(5)}=0$), 
for which a CATE-based inference cannot provide sufficient insight in the causality.
The visualization can be found in Figure \ref{fig:add}. Note that we add 
a standard kernel smoothing since conditional density is assumed to exist in this case. It is clear that both $L_{\mathrm{intra}}^2$-WRF and $L_{\mathrm{inter}}^2$-WRF can provide a good approximation of both $\pi_0(x_{\ast},\cdot)$ and $\pi_1(x_{\ast},\cdot)$.
A more detailed benchmark can be found in Table \ref{tab:supp}. The setting of the experiment (for all considered forests) 
remains the same as in the main text: The dataset is of size 1000 and the associated parameters for the forests are $a_n = 500$, $M=200$ and \textbf{nodesize} $=2$.

 \begin{table}[htb]
  \caption{Estimation of $\pi_0$ (i.e., $\mathcal{L}\left(Y(0)\given X=x\right)$) and $\pi_1$ (i.e., $\mathcal{L}\left(Y(0)\given X=x\right)$)} 
  \label{tab:supp}
  \centering
  \begin{tabular}{lllll}
    \toprule
    Methods  & $\pi_0\text{-}\overline{\mathscr{W}}_1(1000)$ 
    &$\pi_0\text{-}\overline{\mathscr{W}}_2(1000)$
    &$\pi_1\text{-}\overline{\mathscr{W}}_1(1000)$
    &$\pi_1\text{-}\overline{\mathscr{W}}_2(1000)$
    \\
    \midrule
        $L_{\mathrm{intra}}^2$-WRF      &0.6967 &0.8523 & 1.5406 &2.2493\\
        $L_{\mathrm{inter}}^2$-WRF      &\textbf{0.6869} &0.8403 &\textbf{1.3844} &\textbf{1.9881} \\ 
        $L_{\mathrm{inter}}^1$-WRF      &0.6915 &\textbf{0.8397} & 1.4210 &2.0428\\ 
        MF      &2.0110 &2.0321 & 2.3958 & 2.8991\\
        ERT      &0.7025 &0.8961& 1.6490 &2.4223\\ 
        RFCDE     &0.7979 &3.1471 &0.9503&3.3630 \\ 
    \bottomrule
  \end{tabular}
\end{table}
It is clear that $L_{\mathrm{inter}}^2$-WRF provides the overall most accurate prediction for this synthetic dataset. The difference between intra-class and inter-class WRF are more noticeable in the estimation of $\pi_1$, which provides more evidence that inter-class
variants of WRF are better suited for more complex situation (multimodality or large variance).
The fact that $L_{\mathrm{inter}}^2$-WRF outperforms $L_{\mathrm{inter}}^1$-WRF may be due to the existence of conditional density
functions. This case can be regarded as more ``smooth'' than the case considered in the main text, where conditional density does not exist
for $\pi_1$.

\section{On the parameter tuning of WRF}

We discuss in this section the influence of the choice of parameters
(i.e., \textbf{mtry}, $a_n$ and \textbf{nodesize}) of the
WRF and try to provide some suggestions on the algorithm tuning.
We stick to the model provided in Section 3.2 of the main
text and compare the $\pi_t\text{-}\overline{\mathscr{W}}_p(5000)$ respectively
for $t\in \{0,1\}$ and $p\in \{1,2\}$ to illustrate the performance of our method in unimodal
and multimodal situations.
Unlike the conditional expectation estimation, the cross validation-based
tuning strategy is not straightforward to implement
for conditional distribution estimation. 
Indeed, we have
only a single sample at each point $X_i$, and it does not provide enough information
for the conditional distribution.
Therefore, we also track the performance of
the associated conditional
expectation estimations 
in terms of  
Mean Squared Error (MSE).
The conditional expectation functions given $X=x$ of $Y(0)$ and $Y(1)$ are denoted respectively by $\mu_0(x)$ and $\mu_1(x)$.
Our goal is to illustrate whether the tuning for the
conditional expectation can be exploited to guide the tuning for the conditional
distribution estimation problem. 
We also note that since each tree is
constructed using only part of the data, the \emph{out-of-bag} errors for the
forest can thus 
be obtained by averaging the empirical error of each tree on the unused
sub-dataset (see, e.g., \citealp[Section 2.4]{BS15}) in the case where an
independent test dataset is not available.

First, it is well-known that in the classical RF context the number of trees $M$ should be
taken as large as possible, according to the available computing budget,
in order to reduce the variance of the forest.
Although the goal in the WRF framework is changed to the conditional distribution
estimation, it is still suggested to use a large $M$ if possible. 

\begin{figure}[htb]
\centering
\begin{tabular}{ccc}
\subf{\includegraphics[width=36mm]{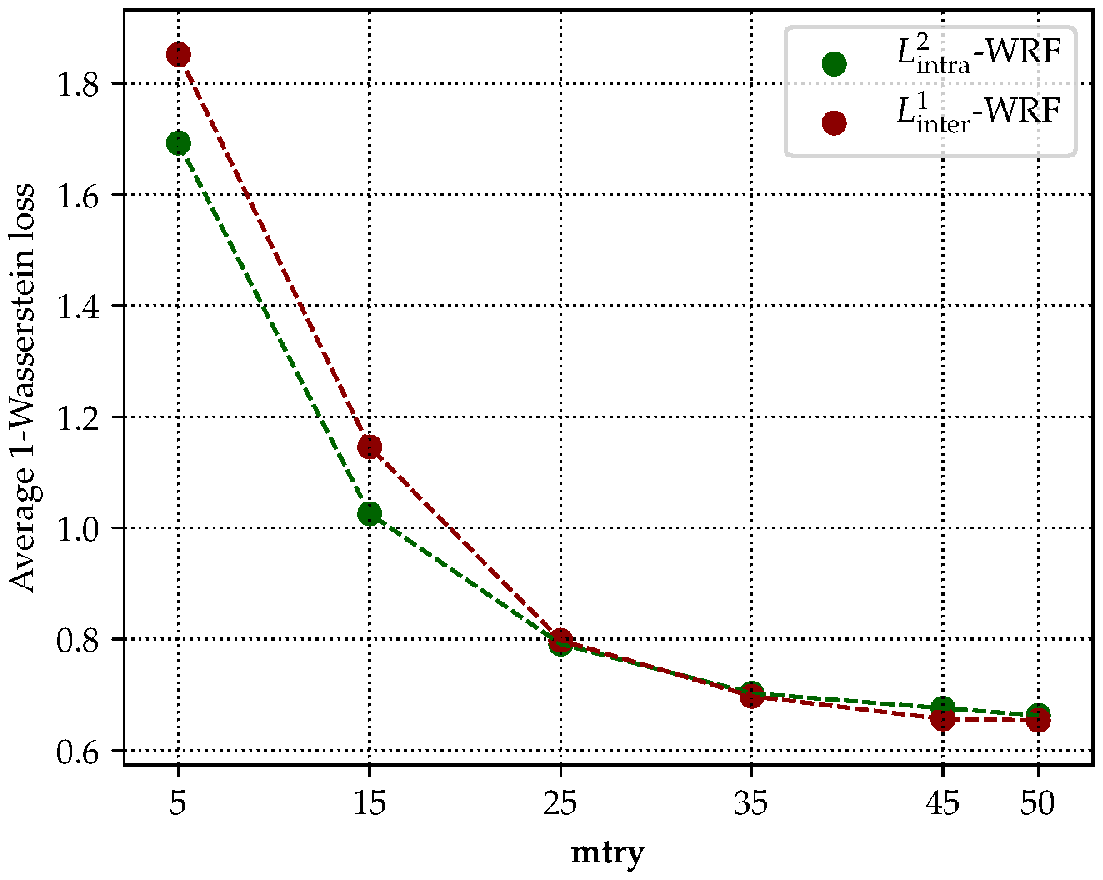}}
{(a) Comparison of $\pi_0$-$\overline{\mathscr{W}}_1(5000)$.}
&
\subf{\includegraphics[width=36mm]{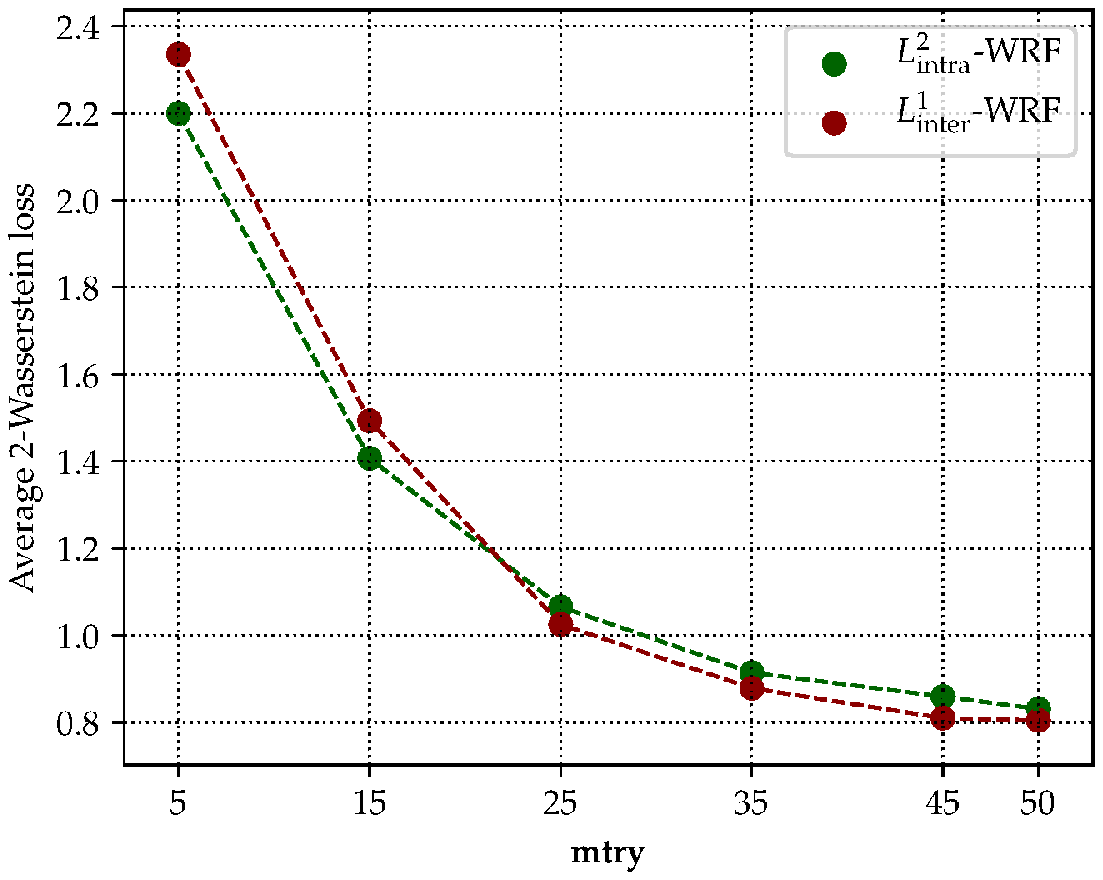}}
{(b) Comparison of $\pi_0$-$\overline{\mathscr{W}}_2(5000)$.}
&
\subf{\includegraphics[width=36mm]{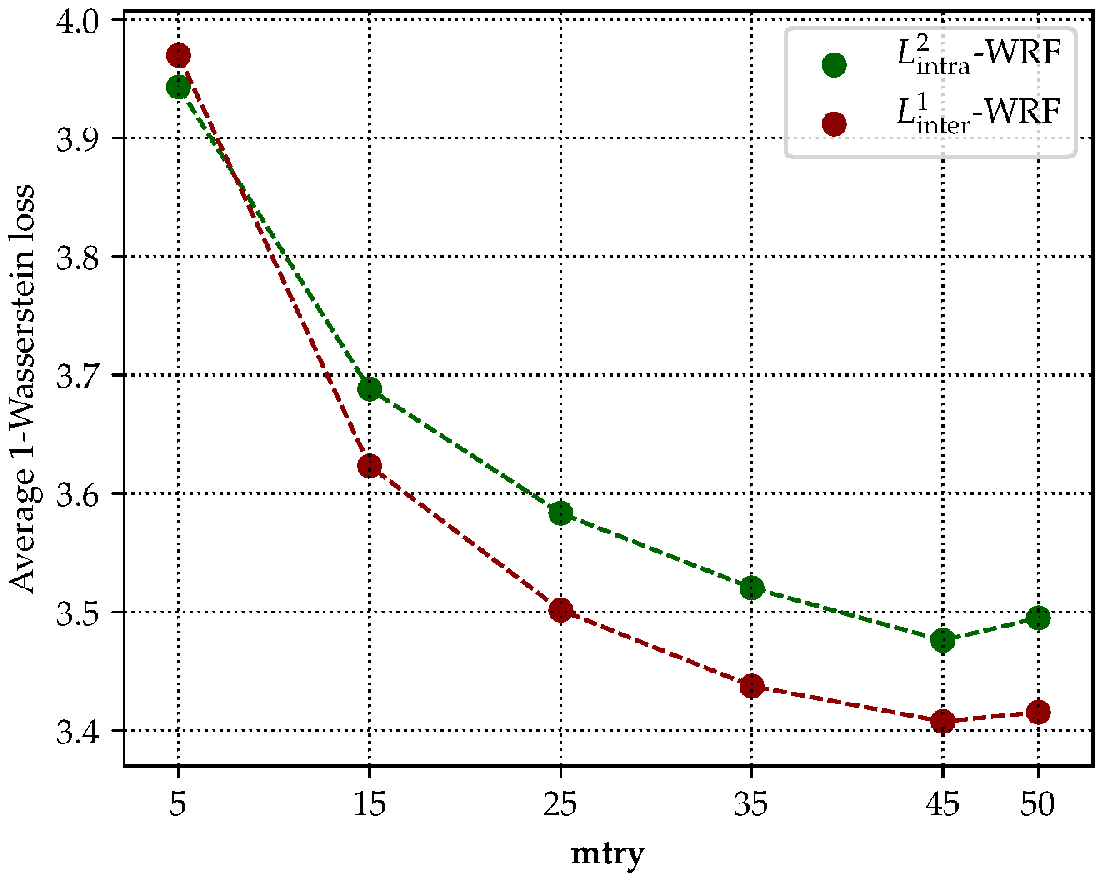}}
{(c) Comparison of $\pi_1$-$\overline{\mathscr{W}}_1(5000)$.}
\\
\subf{\includegraphics[width=36mm]{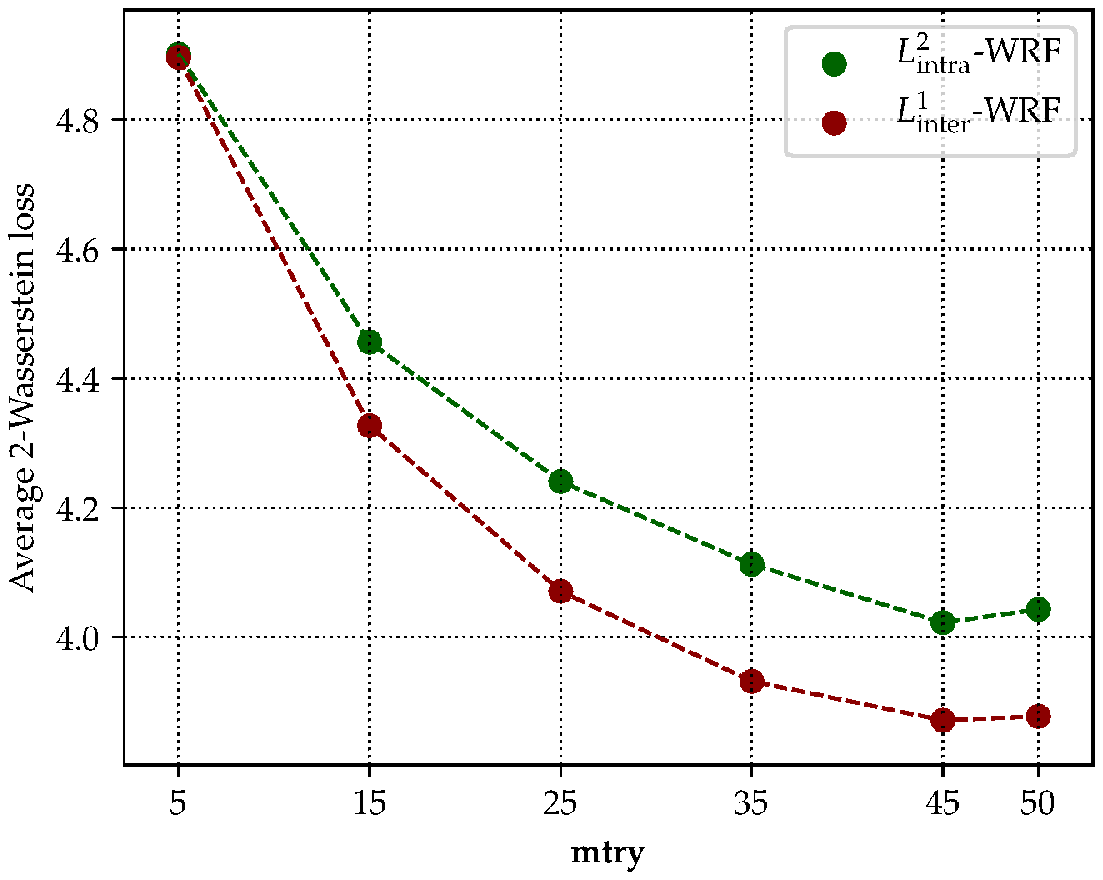}}
{(d) Comparison of $\pi_1$-$\overline{\mathscr{W}}_2(5000)$.}
&
\subf{\includegraphics[width=36mm]{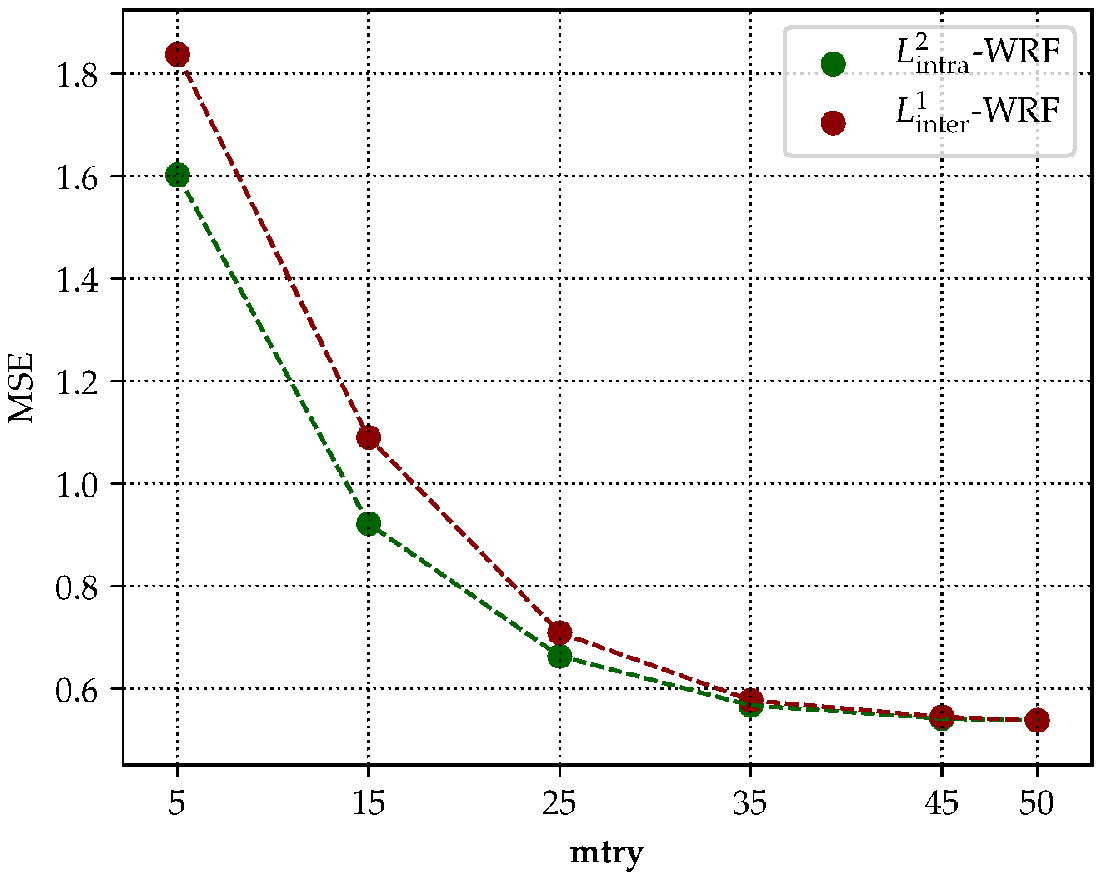}}
{(e) Comparison of the estimation of $\mu_0$.}
&
\subf{\includegraphics[width=36mm]{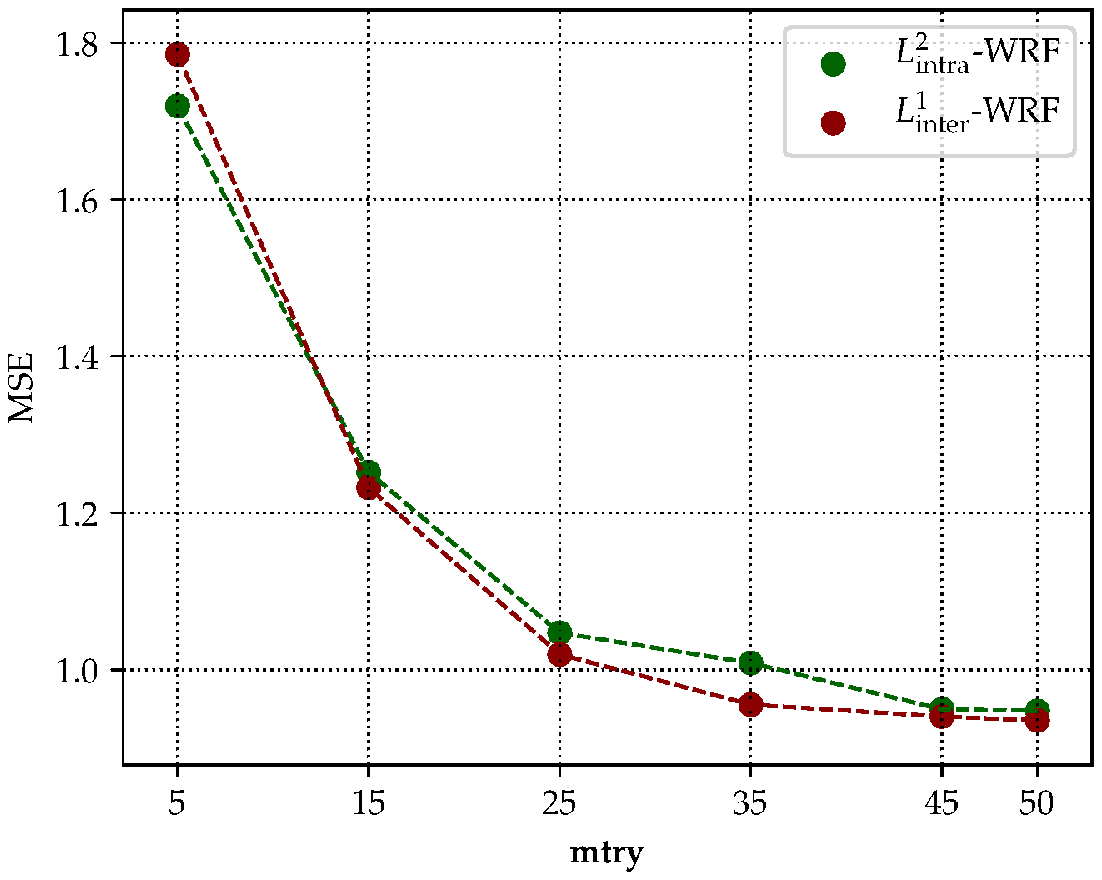}}
{(f) Comparison of the estimation of $\mu_1$.}
\end{tabular}
\caption{An illustration of the performance of different variants of WRF (namely,
  $L_{\mathrm{intra}}^2$-WRF and $L_{\mathrm{inter}}^1$-WRF) with
  $\textbf{mtry}$ varying in $\{5,15,25,35,45,50\}$,  
$a_n = 500$ (\text{with repetition}), $M=300$ and $\textbf{nodesize}=3$.
}
\label{fig:mtry}
\end{figure}

Second, let us investigate the number of directions to be explored at each
cell \textbf{mtry}. 
The result is illustrated in Figure \ref{fig:mtry} ((a)-(d) for average
Wasserstein loss and (e)-(f) for MSE of conditional expectation estimation).
Roughly speaking, the value of \textbf{mtry} reflects the strength of greedy
optimization at each cell during the construction of decision trees. A
conservative approach is to choose \textbf{mtry} as large as possible according
to the available computing resources.

\begin{figure}[tb]
\centering
\begin{tabular}{ccc}
\subf{\includegraphics[width=36mm]{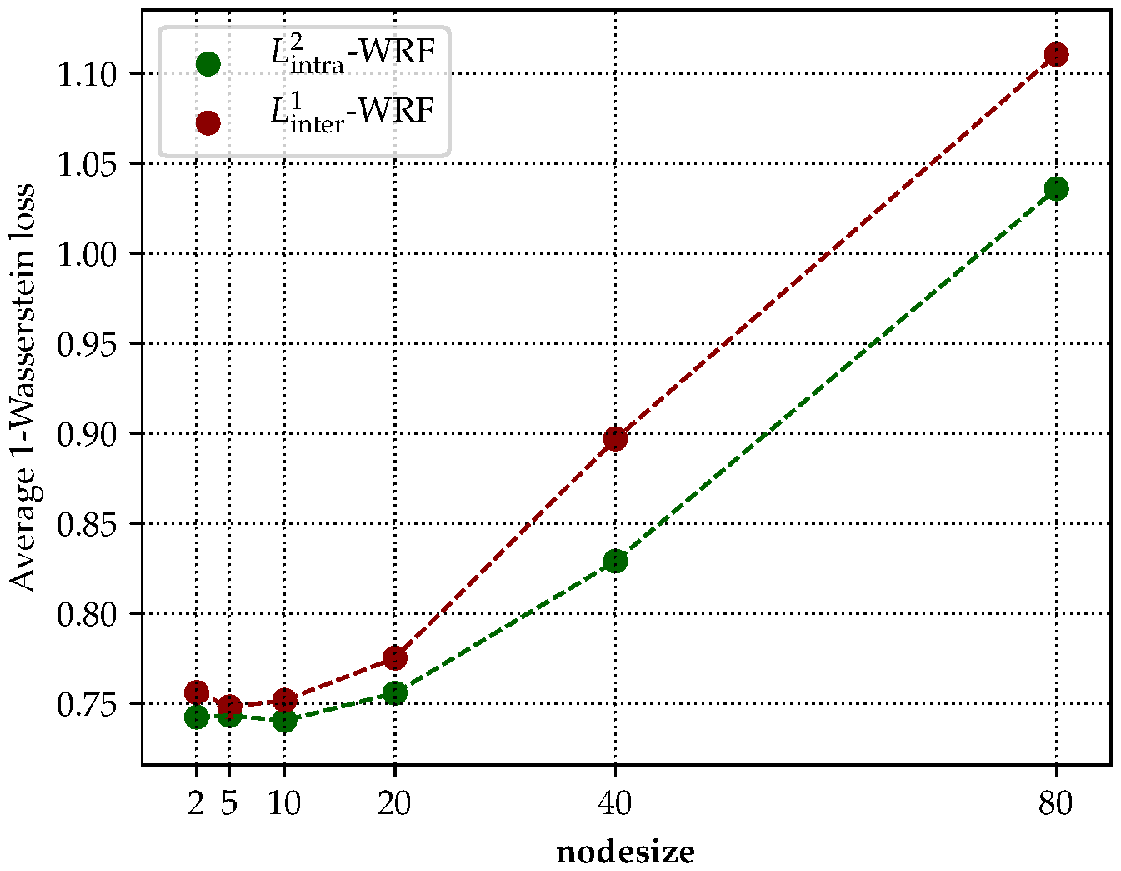}}
{(a) Comparison of $\pi_0$-$\overline{\mathscr{W}}_1(5000)$.}
&
\subf{\includegraphics[width=36mm]{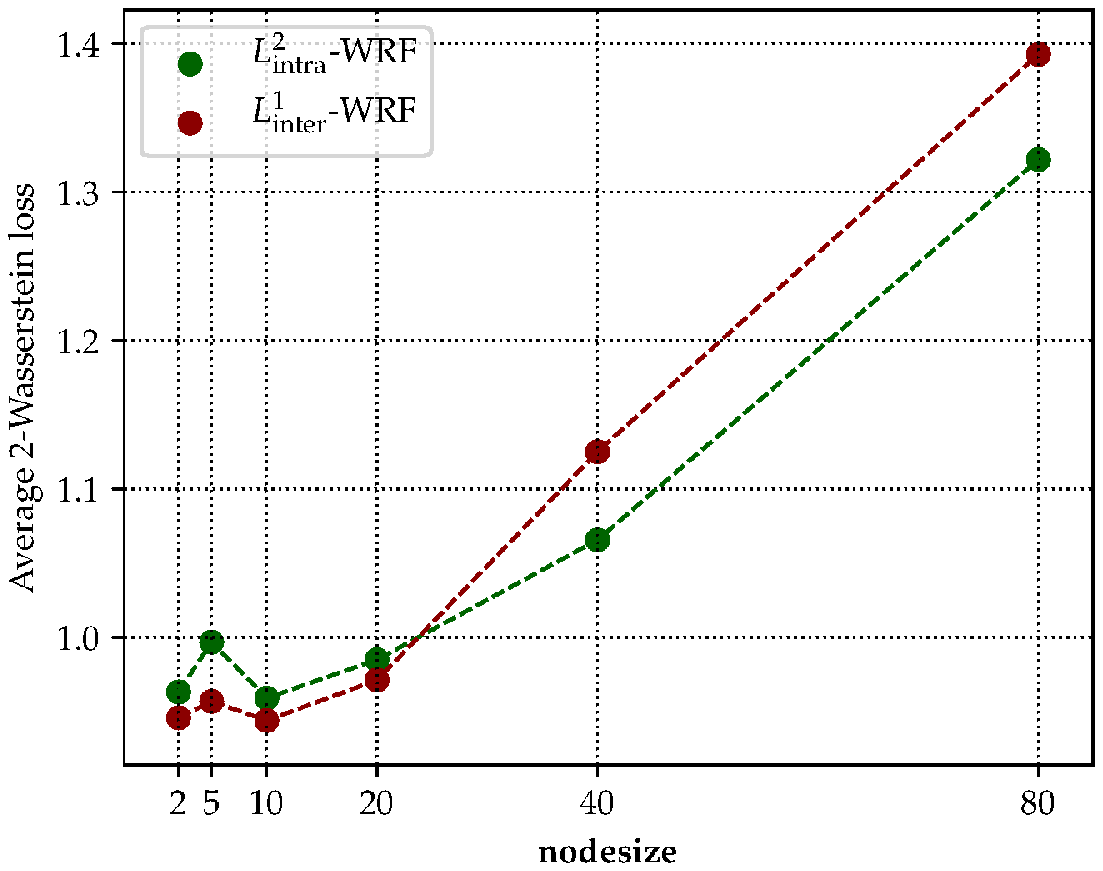}}
{(b) Comparison of $\pi_0$-$\overline{\mathscr{W}}_2(5000)$.}
&
\subf{\includegraphics[width=36mm]{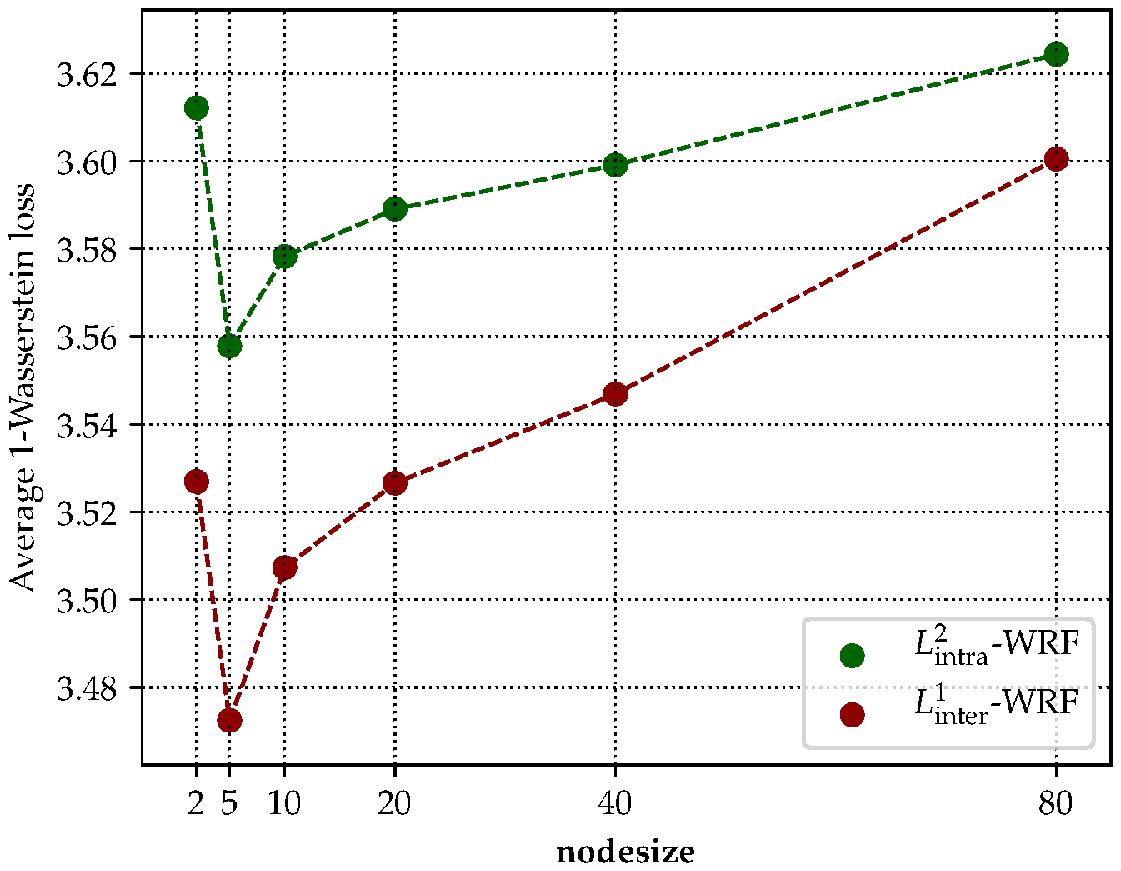}}
{(c) Comparison of $\pi_1$-$\overline{\mathscr{W}}_1(5000)$.}
\\
\subf{\includegraphics[width=36mm]{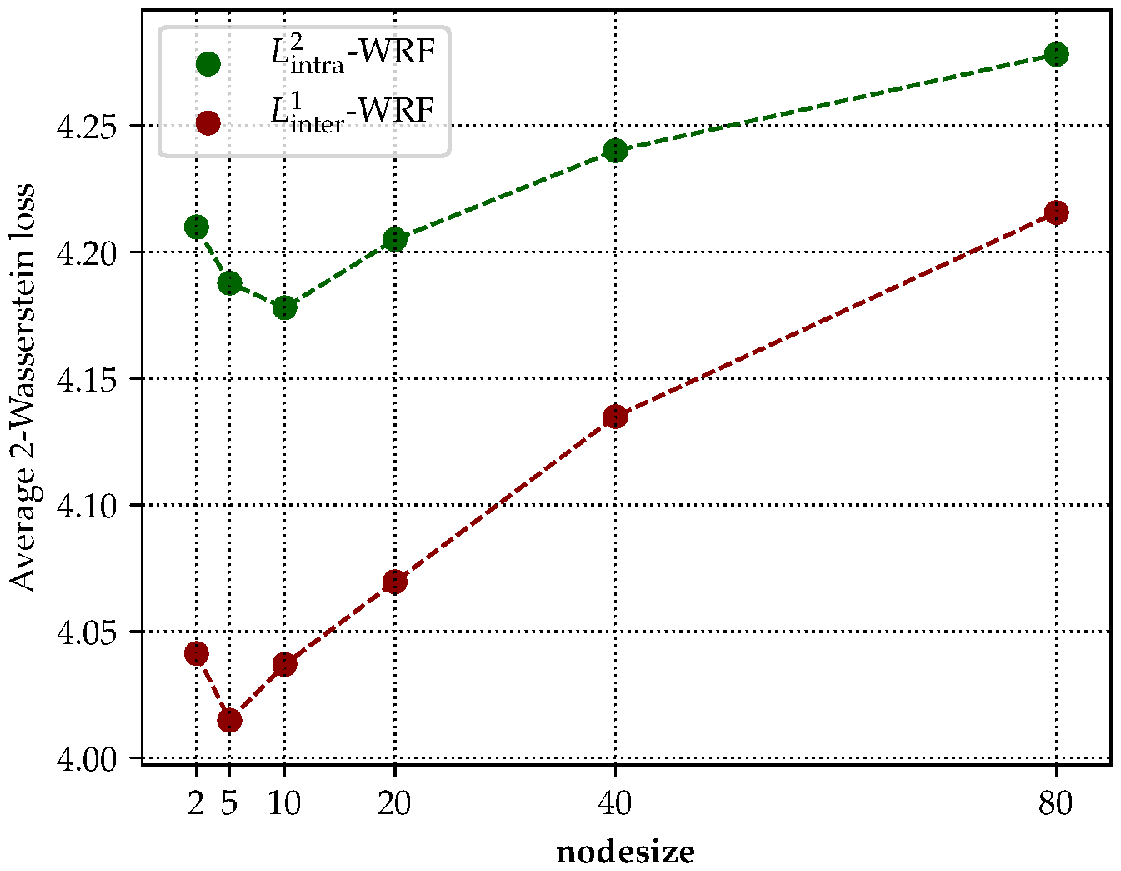}}
{(d) Comparison of $\pi_1$-$\overline{\mathscr{W}}_2(5000)$.}
&
\subf{\includegraphics[width=36mm]{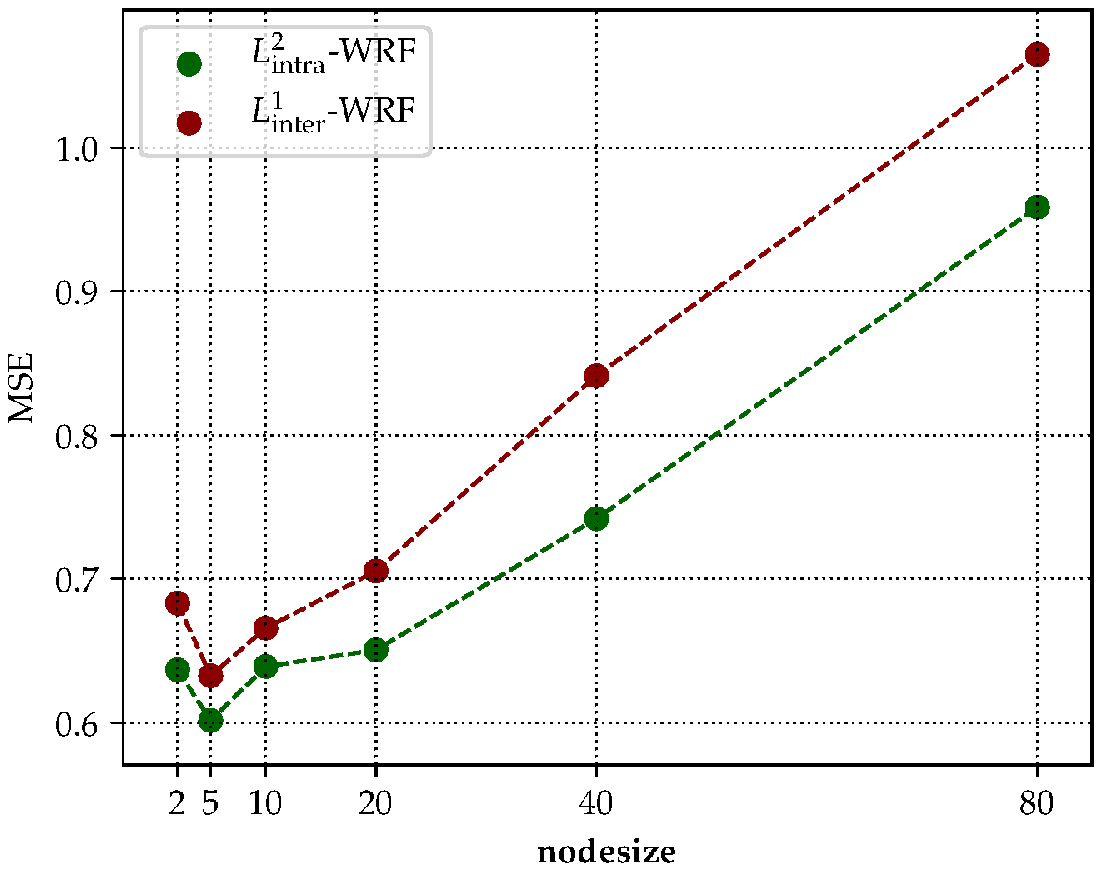}}
{(e) Comparison of the estimation of $\mu_0$.}
&
\subf{\includegraphics[width=36mm]{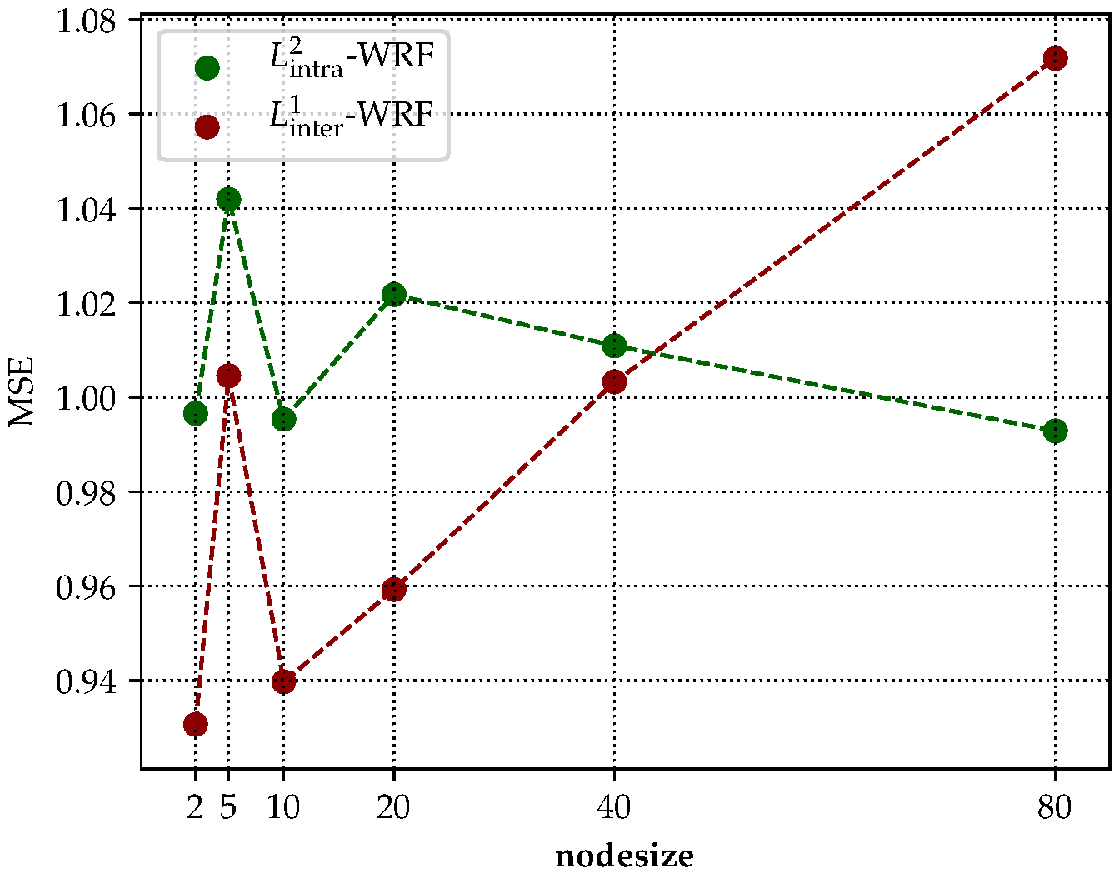}}
{(f) Comparison of the estimation of $\mu_1$.}
\end{tabular}
\caption{An illustration of the performance of different variants of WRF (namely,
  $L_{\mathrm{intra}}^2$-WRF and $L_{\mathrm{inter}}^1$-WRF) with
  $\textbf{nodesize}$ varying in $\{2,5,10,20,40,80\}$,
$a_n = 500$ (\text{with repetition}), $M=300$ and $\textbf{mtry}=30$.
}
\label{fig:nodesize}
\end{figure}

Then, let us see the influence brought by the change of \textbf{nodesize}. The
illustration can be found in Figure \ref{fig:nodesize} ((a)-(d) for average
Wasserstein loss and (e)-(f) for MSE of conditional expectation estimation).
In the classical RF context, the motivation of the choice $\textbf{nodesize}
>2$ can be interpreted as introducing some local averaging procedure at each cell
in order to deal with
the variance or noise of the sample. 
Here, as discussed in the main
text, we are interested in the conditional distribution estimation
in the HTE context, where the variance or other fluctuation of the
conditional distribution is part of the information to be estimated. Hence, the
interpretation of the choice $\textbf{nodesize}>2$ should be adapted
accordingly, as the
minimum sample size that is used to describe the conditional distribution
at each cell. This interpretation is better suited when it comes to the
estimation of multimodal conditional distributions. 
As shown in Figure \ref{fig:nodesize} (a)-(d), there are some
optimal choices of \textbf{nodesize} between $2$ and $a_n$. In the simple cases,
such as the estimation of $\pi_0$ (unimodal), the MSE of the associated conditional
expectation (Figure \ref{fig:nodesize} (e)) can be used, accordingly, to tune the algorithm
for conditional distribution estimation. However, in the more complex case such as
the estimation of $\pi_1$ (bi-modal), the MSE of the conditional expectation
estimation is no as stable (Figure \ref{fig:nodesize} (f)). Nevertheless, it is
also recommended to use small \textbf{nodesize} in this situation as a conservative choice.

\begin{figure}[htb]
\centering
\begin{tabular}{ccc}
\subf{\includegraphics[width=36mm]{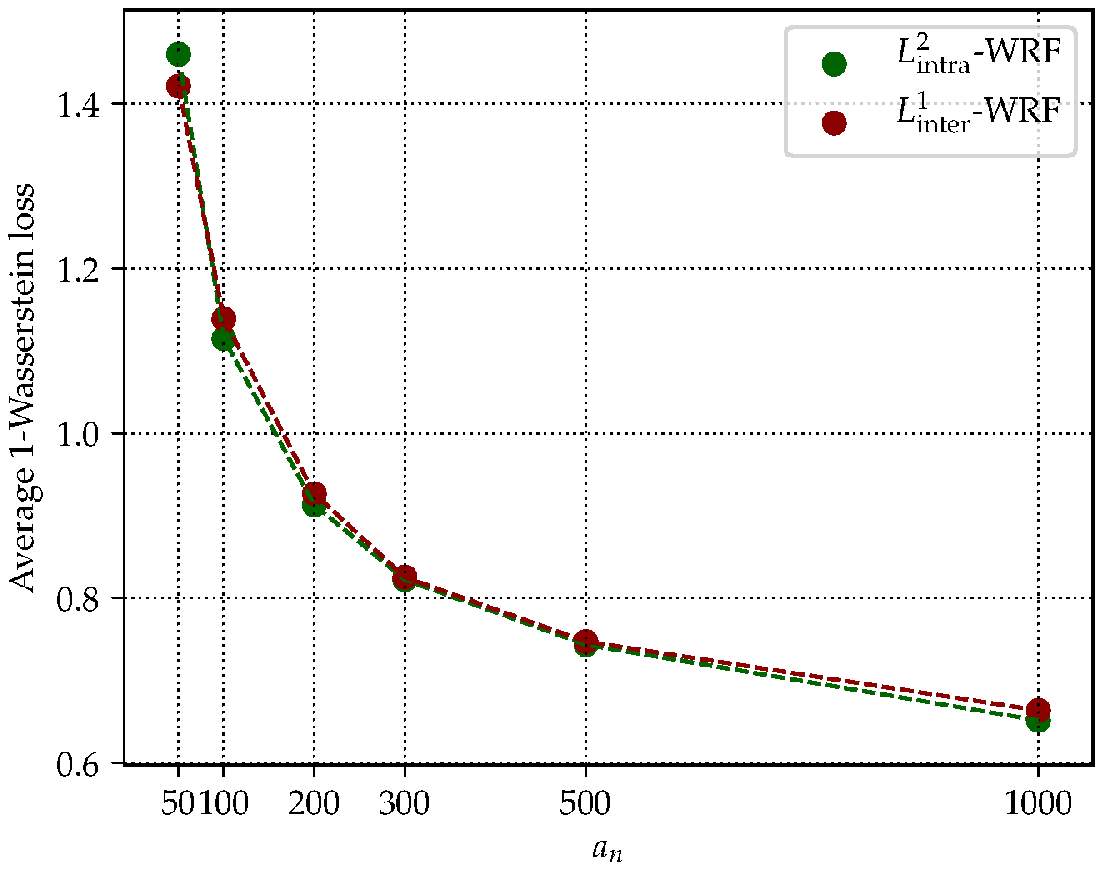}}
{(a) Comparison of $\pi_0$-$\overline{\mathscr{W}}_1(5000)$.}
&
\subf{\includegraphics[width=36mm]{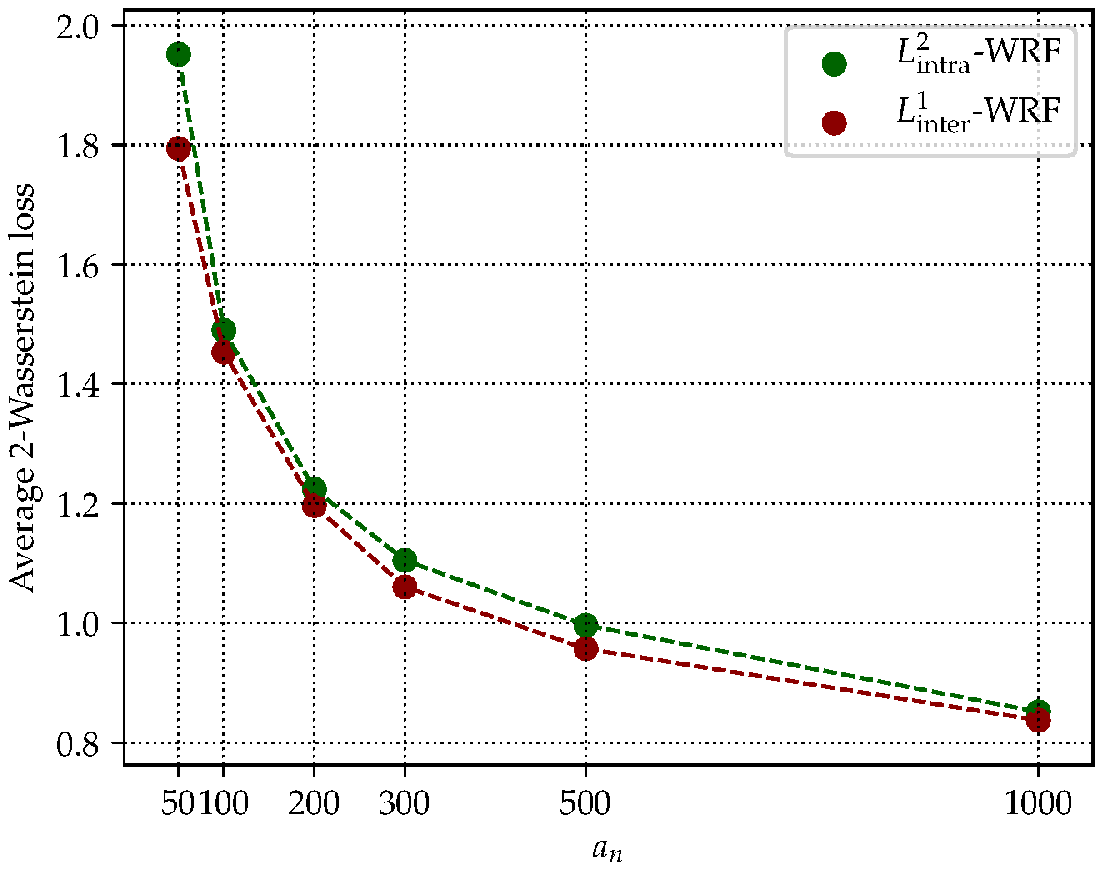}}
{(b) Comparison of $\pi_0$-$\overline{\mathscr{W}}_2(5000)$.}
&
\subf{\includegraphics[width=36mm]{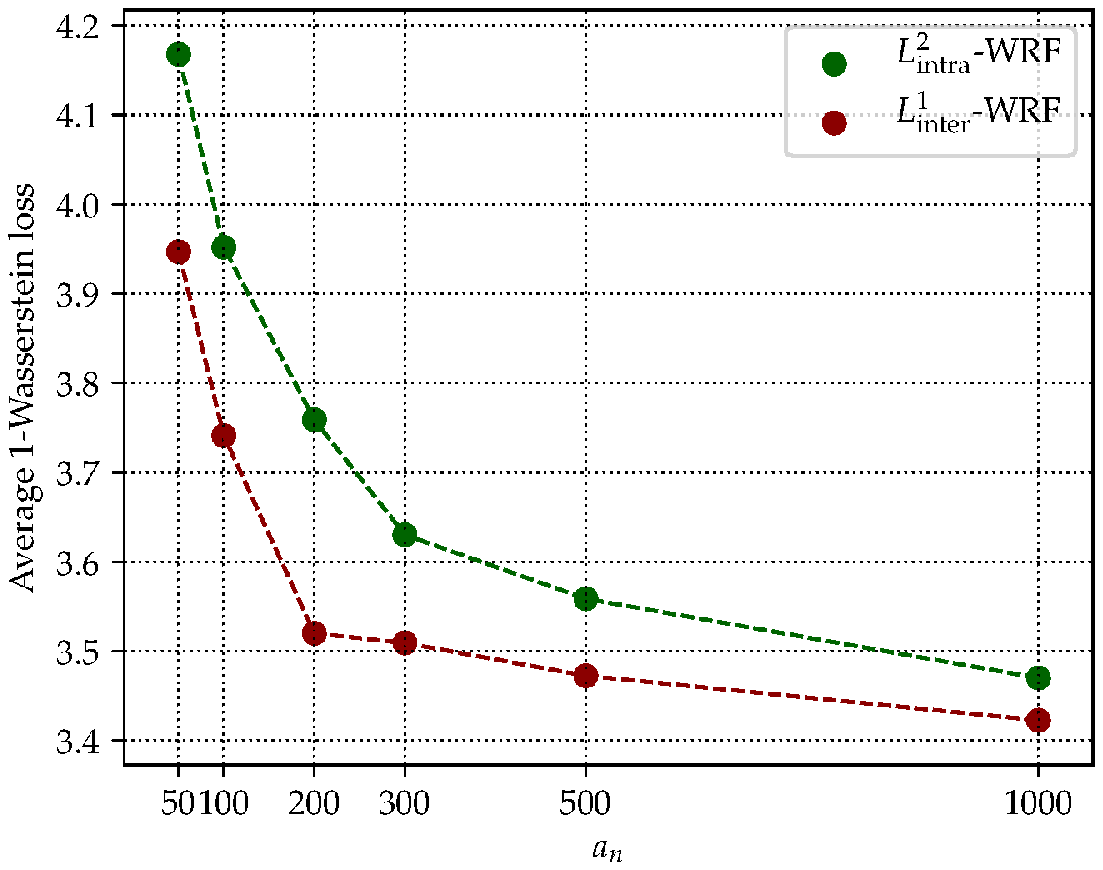}}
{(c) Comparison of $\pi_1$-$\overline{\mathscr{W}}_1(5000)$.}
\\
\subf{\includegraphics[width=36mm]{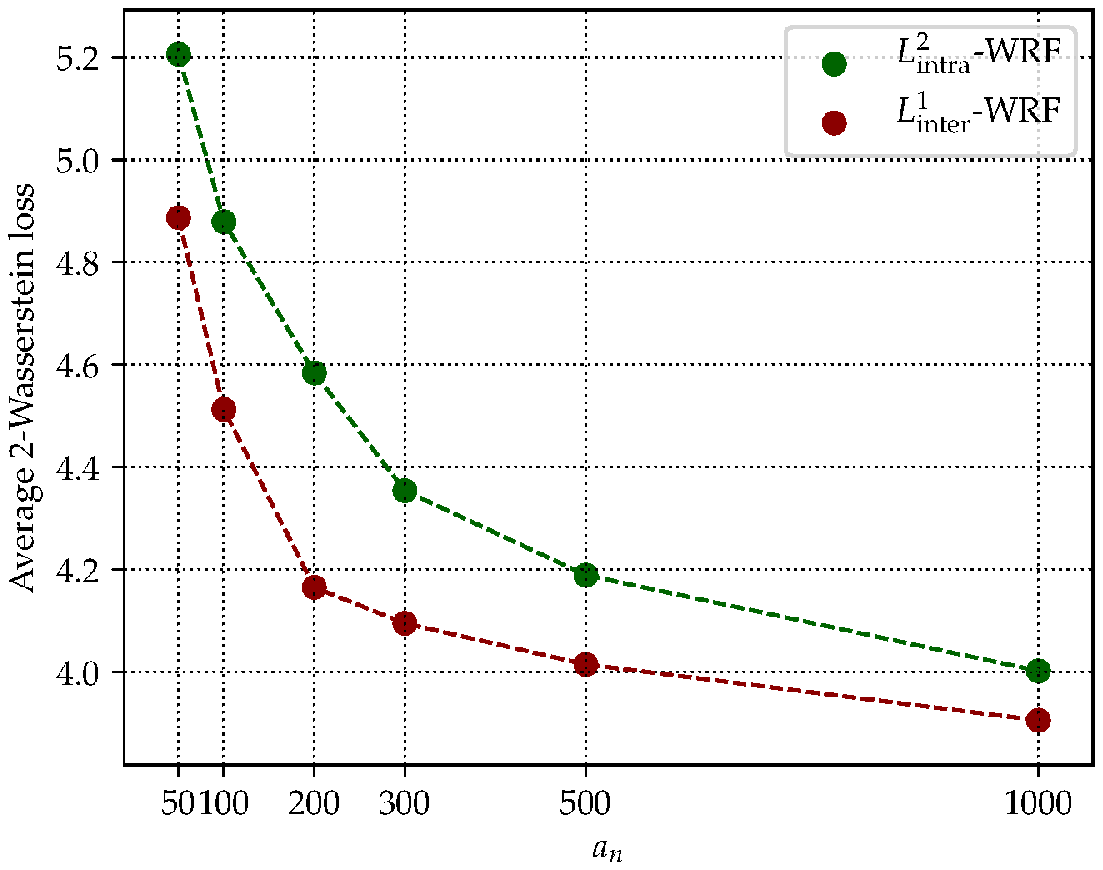}}
{(d) Comparison of $\pi_1$-$\overline{\mathscr{W}}_2(5000)$.}
&
\subf{\includegraphics[width=36mm]{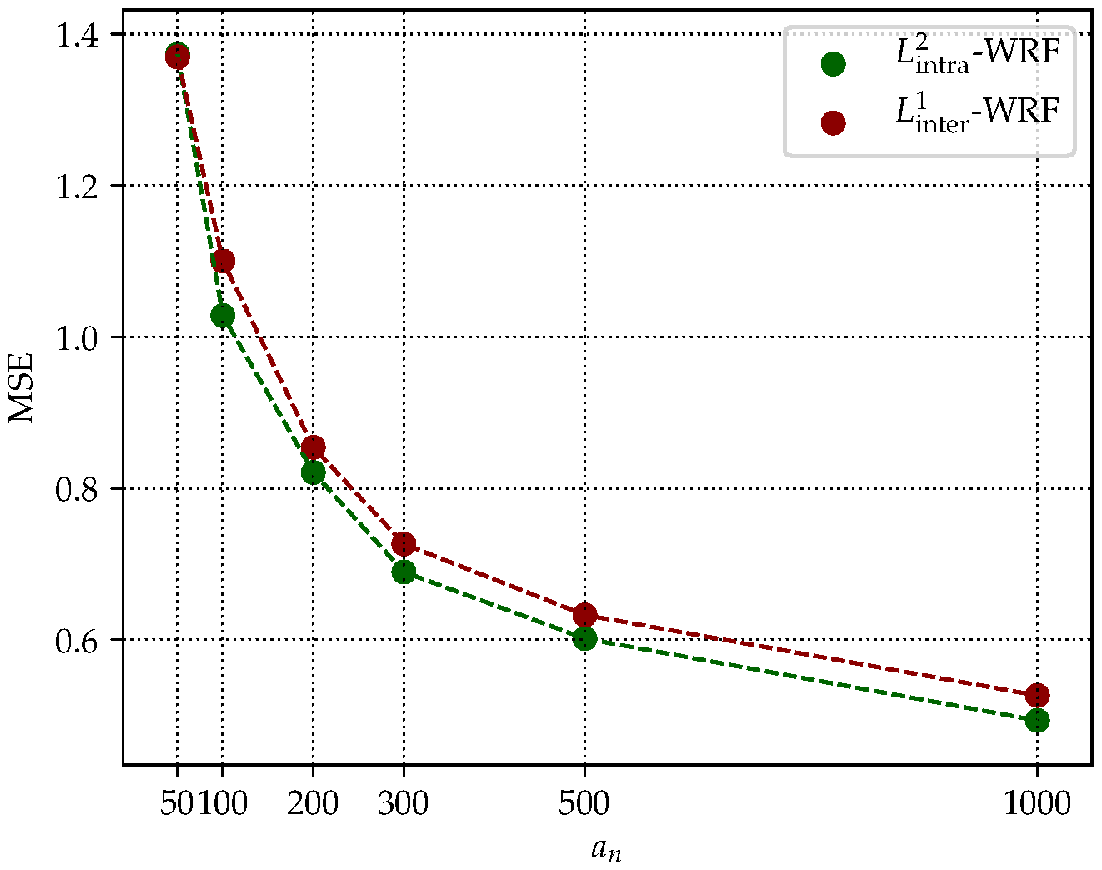}}
{(e) Comparison of the estimation of $\mu_0$.}
&
\subf{\includegraphics[width=36mm]{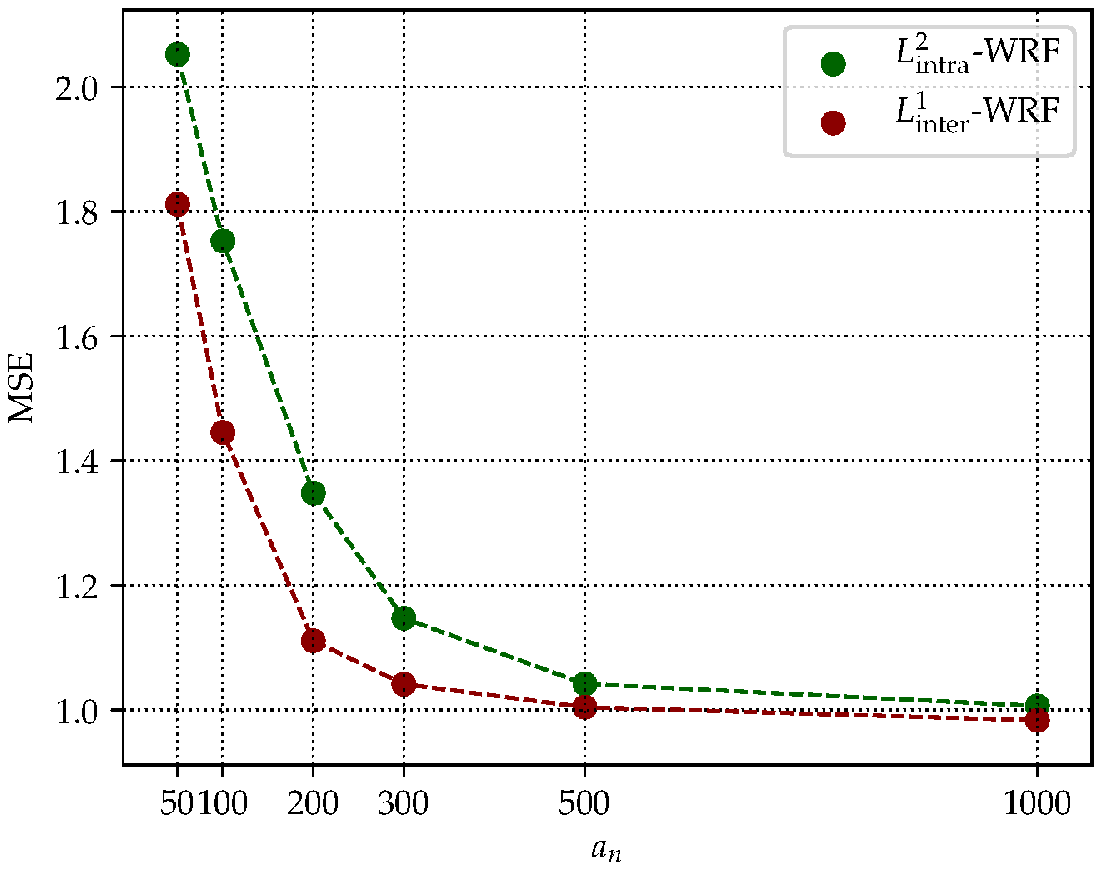}}
{(f) Comparison of the estimation of $\mu_1$.}
\end{tabular}
\caption{An illustration of the performance of different variants of WRF (namely,
  $L_{\mathrm{intra}}^2$-WRF and $L_{\mathrm{inter}}^1$-WRF) with
  $a_n$ varies in $\{50,100,200,300,500,1000\}$ (\text{with repetition}),
  $\textbf{nodesize} = 5$, $M=300$ and $\textbf{mtry}=30$.
}
\label{fig:a_n}
\end{figure}

Finally, we discuss the size $a_n$ of the sub-dataset used to construct each
decision tree. 
Note that the choice of $a_n$ is still not well-understood even in
the classical RF context (see, e.g., \citealp{BS15,consistency-RF}).  
When the 
computing budget allows to implement $a_n = n$ (with replacement, which
corresponds to the classical Bootstrap), we recommend
to use this choice. Otherwise, we recommend to fix the $a_n$ from one fifth to one third of the whole data size in
order to maintain a reasonably good performance without heavy computations.

\paragraph{Suggestions on the parameter tuning} 
The take-home message for the parameter tuning of WRF is simple:
We recommend
to use large $M$ and \textbf{mtry} according to the available
computing resources. The parameter \textbf{nodesize} can be tuned via
a cross validation-based
strategy using the MSE of the associated conditional expectation estimation.
In addition, we suggest to choose smaller \textbf{nodesize} when there is abnormal
fluctuation of the MSE score. 
It is also proposed to use classical bootstrap (i.e.,
$a_n = n$ with replacement) when possible. Otherwise, we suggest to fix
a smaller $a_n$ according to the computing budget. 
Finally, although there is no theoretical guarantee, we advocate to use
$L_{\mathrm{inter}}^1$-WRF or $L_{\mathrm{inter}}^2$-WRF for univariate objective, since it has a better
overall accuracy with a reasonable additional computational cost.

\section{On the propensity score function}
\begin{figure}[htb]
\centering
\begin{tabular}{cc}
\subf{\includegraphics[width=50mm]{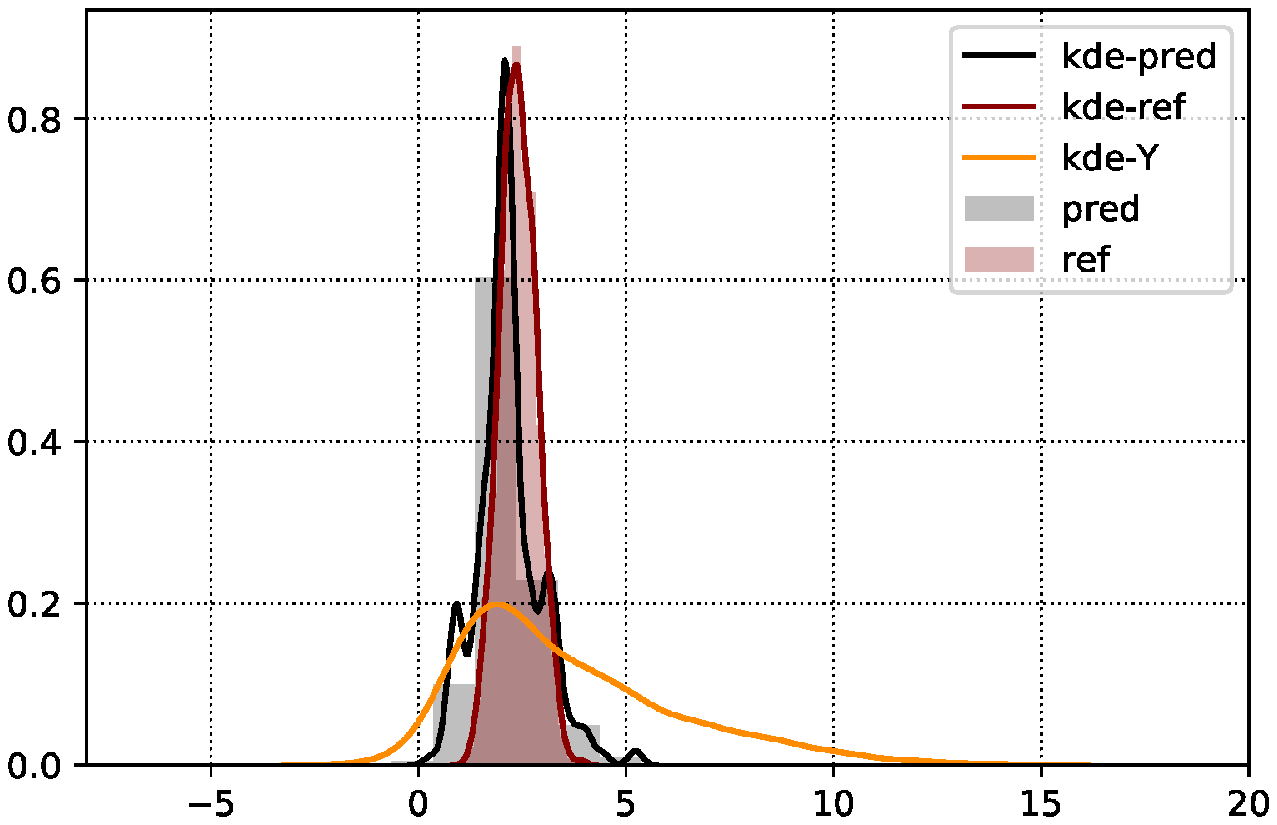}}
     {(a) $\pi_0(x_{\ast},\cdot)$ estimated by $L_{\mathrm{intra}}^2$-WRF}
&
\subf{\includegraphics[width=50mm]{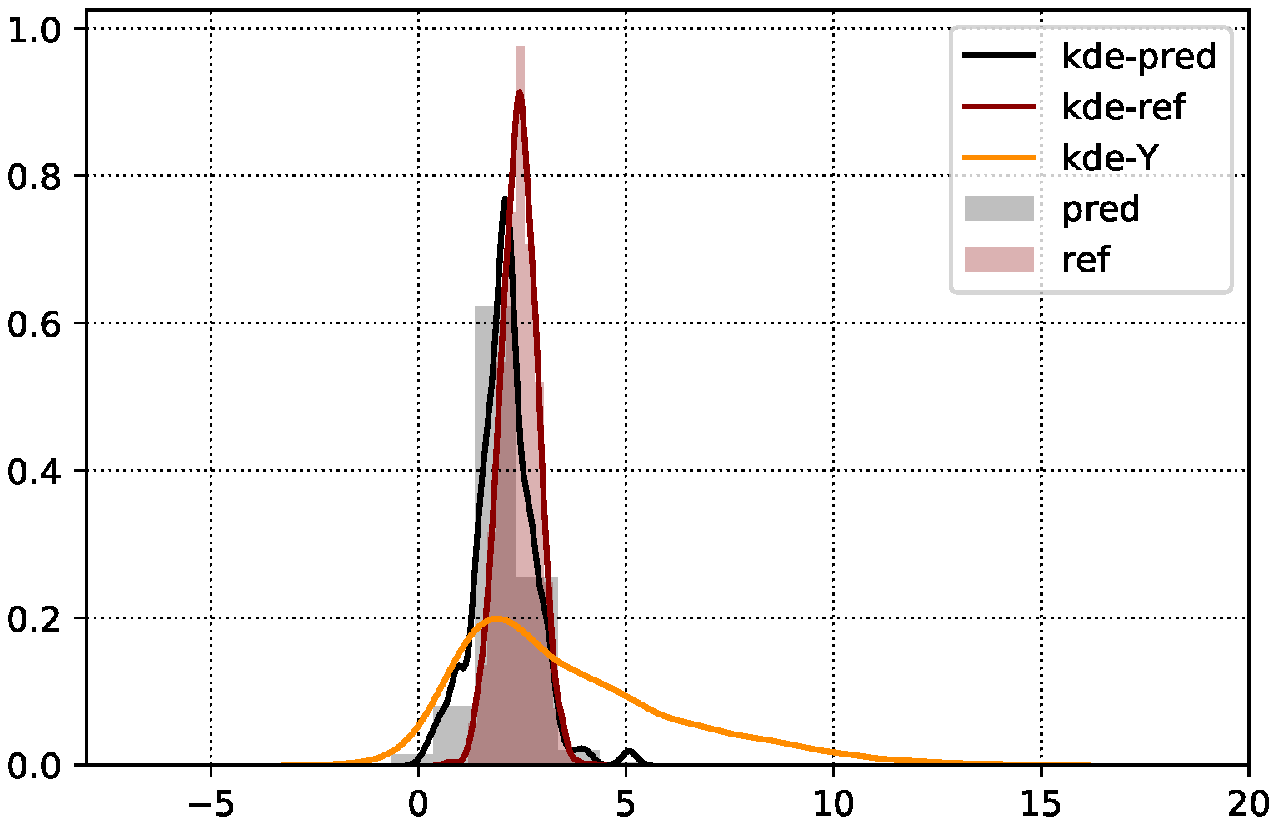}}
     {(b) $\pi_0(x_{\ast},\cdot)$ estimated by $L_{\mathrm{inter}}^1$-WRF}
\\
\subf{\includegraphics[width=50mm]{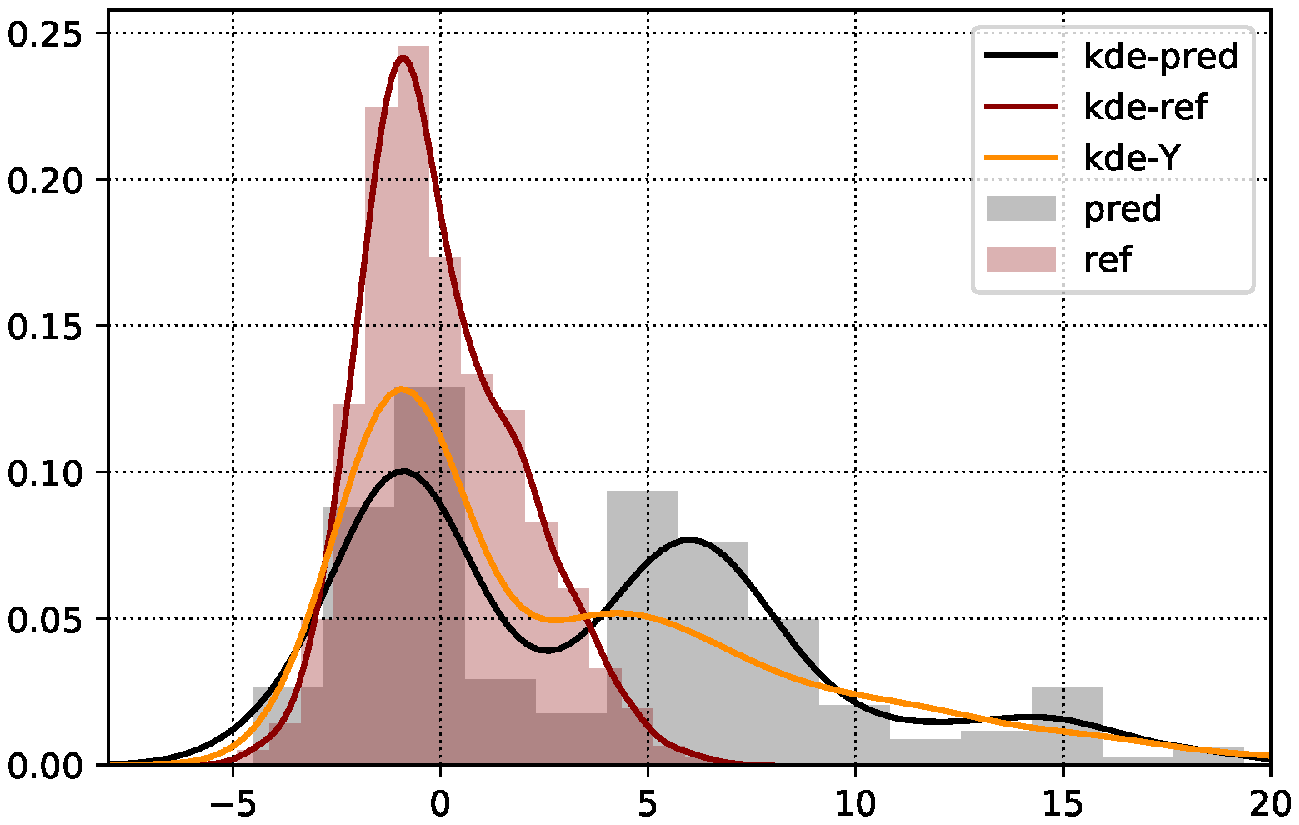}}
     {(c) $\pi_1(x_{\ast},\cdot)$ estimated by $L_{\mathrm{intra}}^2$-WRF}
&
\subf{\includegraphics[width=50mm]{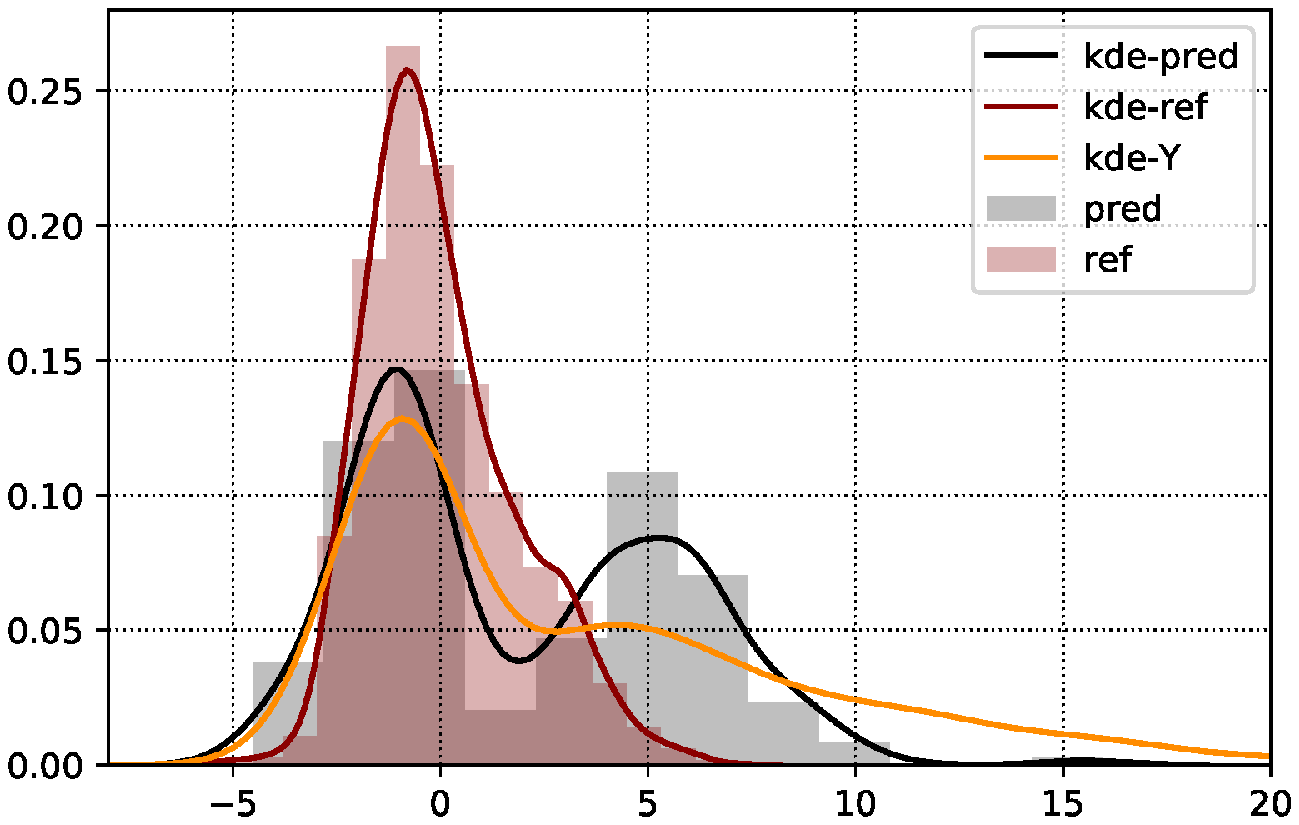}}
     {(d) $\pi_1(x_{\ast},\cdot)$ estimated by $L_{\mathrm{inter}}^1$-WRF}
\end{tabular}
\caption{An illustration of estimated conditional distributions provided 
by different variants of WRF with the same parameters: 
$a_n = 500$ (\text{with repetition}), $M=200$, $\textbf{mtry} = 50$, $\textbf{nodesize}=2$.
In the legend, \texttt{pred} and \texttt{ref} denote respectively the prediction given by WRF and reference values sampled directly from the true conditional distribution with sample size fixed to be 2000.
The acronyms \texttt{kde}-\texttt{pred} and \texttt{kde}-\texttt{ref} stand for the outputs of the \texttt{kdeplot} function of \texttt{seaborn} package \citep{seaborn}, which provides a standard kernel smoothing. Finally, \texttt{kde}-\texttt{Y} denotes the \texttt{kdeplot} of the $Y$-population, i.e., all the $Y_i(1)$ or $Y_i(0)$ in the training dataset according to the treatment/control group. 
}
\label{fig:ps}
\end{figure}

The propensity score function $e(\cdot)$ measures the probability that the treatment is
assigned to a certain individual, which basically determines the distribution
of the available
dataset for the estimation of $\pi_0$ and $\pi_1$ in the population. 
More precisely, 
imagine that $x$ is an individual such that in the neighbourhood of
$x$, the value of $e(\cdot)$ is close to 0. Then, it is expected that only very few
training data for the estimation of $\pi_1(x,\cdot)$ can be collected during the
observational study. As a consequence, it is expected that the estimation
$\hat{\pi}_1$ at such point is of reasonably bad quality. 
For example,
the propensity score function is
\[
  e(x) = \frac{1}{2} \sin(2x^{(1)}x^{(2)}+6x^{(3)}) + \frac{1}{2}.
\]
Denote by $x_{\star}$ an individual such that
$x_{\star}^{(1)} = \frac{\pi}{4}$, $x_{\star}^{(2)} = 1$, and $x_{\star}^{(6)} = \frac{\pi}{6}$. 
It is readily checked that $e(x_{\star}) = 0$. As shown in Figure \ref{fig:ps},
the estimation of $\pi_0(x_{\star},\cdot)$ is very accurate (see Figure \ref{fig:ps} (a)-(b)),
while the estimation of $\pi_1(x_{\star},\cdot)$ is of poor quality (see Figure
\ref{fig:ps} (c)-(d)).

From a theoretical
perspective, one may suppose that the propensity score function is bounded away
from 0 and 1 uniformly for all $x\in \mathbb{R}^d$ (see, e.g.,
\citealp{KSBY17,NW17}). However, it is, unfortunately, not possible to
control the propensity score during an observational study. 
As a consequence, it is usually very difficult to verify such an assumption in practice. 
Therefore, a more meaningful question can be how to detect if our
estimation is reliable or not for a certain individual.
A straightforward strategy is to estimate the
propensity score function independently, as done for example in \citep{AW19}, and to test whether the
value of this score is close to 0 and 1.
Another
approach is to exploit the information encoded in the splits/weights of the
forest to detect whether enough data is collected for the prediction at
target individual. The details are left for future research.

Finally, let us mention that if the goal is to estimate the function $\Lambda_p(\cdot)$ defined
in Section 3.1 of the main text, we expect that more dedicated variants of WRF can be
constructed, in the same spirit of Causal Forests introduced in \citep{AW19}.

\section{Possible extensions}
In this section, we discuss two natural extensions of WRF that we did not investigate
in details. 

First, inspired by the Random Rotation Ensembles
introduced in \citep{RRE}, it is natural to consider the implementation of oblique splits, i.e., the splits
are not necessarily axis-aligned.
More precisely, for each tree, by sampling a
uniformly distributed rotation matrix (e.g.\@ \citealp[Section 3]{RRE}), we are able to construct the decision tree by
using the rotated sub-dataset (or equivalently, one can also implement randomly rotated cuts in the tree's construction). Intuitively speaking, the rotation variants of
WRF will be more consistent when it comes to performance, while the additional
computing resources are required for both training and prediction.

Another direction is to replace the Dirac mass in the empirical
measures by some kernel $K(x,dy)$, as proposed in \citep{RFCDE}. 
For instance, the $L_{\mathrm{inter}}^p$-WRF can be modified by using the
following splitting criteria:
\[
  \begin{aligned}
  \tilde{L}_{\mathrm{inter}}^p(A_L,A_R)
    :=&
  \frac{N_L}{N_A} 
  \mathscr{W}_p^p\left(
    \frac{1}{N_L}\sum_{X_i\in A_L} K(Y_i,dy),
    \frac{1}{N_A}\sum_{X_i\in A} K(Y_i,dy)
  \right)
    \\&+
  \frac{N_R}{N_A}
  \mathscr{W}_p^p\left(
    \frac{1}{N_R}\sum_{X_i\in A_R} K(Y_i,dy),
    \frac{1}{N_A}\sum_{X_i\in A} K(Y_i,dy)
  \right),
  \end{aligned}
\]
where the kernel $K(\cdot,\cdot)$ is chosen according to prior knowledge of the problem. 
At the same time, the final prediction will be replaced by
\[
  \tilde{\pi}_{M,n}(x,dy;\Theta_{[M]},\mathcal{D}_n)
=\sum_{i = 1}^n \alpha_{i}(x) K(Y_i,dy),
\]
where $\alpha_{i}(\cdot)$ remains the same as defined in Section 2.1 of the main text.
When the associated $\mathscr{W}_p$-distance is easy to compute, we expect that
this extension will be more accurate for small datasets.
Nevertheless, the performances of these natural extensions are still not clear.
The details are therefore left for future research.

\end{appendices}

\bibliography{citation}

\begin{thebibliography}{}

\bibitem[Athey et~al., 2019]{ATW19}
Athey, S., Tibshirani, J., and Wager, S. (2019).
\newblock Generalized random forests.
\newblock {\em The Annals of Statistics}, 47:1148--1178.

\bibitem[{Athey} and {Wager}, 2019]{AW19}
{Athey}, S. and {Wager}, S. (2019).
\newblock {Estimating treatment effects with causal forests: An application}.
\newblock {\em Observational Studies}, 5.

\bibitem[Biau et~al., 2015]{biau2015new}
Biau, G., C{\'e}rou, F., and Guyader, A. (2015).
\newblock New insights into approximate bayesian computation.
\newblock {\em Annales de l'IHP Probabilit{\'e}s et statistiques},
  51(1):376--403.

\bibitem[Biau and Scornet, 2015]{BS15}
Biau, G. and Scornet, E. (2015).
\newblock A random forest guided tour.
\newblock {\em TEST}, 25:197--227.

\bibitem[Blaser and Fryzlewicz, 2016]{RRE}
Blaser, R. and Fryzlewicz, P. (2016).
\newblock Random rotation ensembles.
\newblock {\em Journal of Machine Learning Research}, 17:1--26.

\bibitem[Breiman, 2001]{RF}
Breiman, L. (2001).
\newblock Random forests.
\newblock {\em Machine Learning}, 45:5--32.

\bibitem[Chernozhukov et~al., 2018]{Chernozhukov}
Chernozhukov, V., Demirer, M., Duflo, E., and Fernández-Val, I. (2018).
\newblock Generic machine learning inference on heterogeneous treatment effects
  in randomized experiments.
\newblock Working Paper 24678, National Bureau of Economic Research.

\bibitem[Cuturi, 2013]{sinkhorn_nips}
Cuturi, M. (2013).
\newblock Sinkhorn distances: Lightspeed computation of optimal transport.
\newblock In Burges, C., Bottou, L., Welling, M., Ghahramani, Z., and
  Weinberger, K., editors, {\em Advances in Neural Information Processing
  Systems 26}, pages 2292--2300. Curran Associates, Inc.

\bibitem[Dutordoir et~al., 2018]{GPCDE}
Dutordoir, V., Salimbeni, H., Hensman, J., and Deisenroth, M. (2018).
\newblock Gaussian process conditional density estimation.
\newblock In Bengio, S., Wallach, H., Larochelle, H., Grauman, K.,
  Cesa-Bianchi, N., and Garnett, R., editors, {\em Advances in Neural
  Information Processing Systems 31}, pages 2385--2395. Curran Associates, Inc.

\bibitem[Genevay et~al., 2019]{sinkhorn_sample}
Genevay, A., Chizat, L., Bach, F., Cuturi, M., and Peyr\'{e}, G. (2019).
\newblock Sample complexity of {S}inkhorn divergences.
\newblock In Chaudhuri, K. and Sugiyama, M., editors, {\em Proceedings of
  Machine Learning Research}, volume~89 of {\em Proceedings of Machine Learning
  Research}, pages 1574--1583. PMLR.

\bibitem[Geurts et~al., 2006]{ERT}
Geurts, P., Ernst, D., and Wehenkel, L. (2006).
\newblock Extremely randomized trees.
\newblock {\em Machine Learning}, 63:3–42.

\bibitem[Hall et~al., 1999]{CDE1}
Hall, P., Wolff, R.~C., and Yao, Q. (1999).
\newblock Methods for estimating a conditional distribution function.
\newblock {\em Journal of the American Statistical Association}, 94:154--163.

\bibitem[Hall and Yao, 2005]{CDE2}
Hall, P. and Yao, Q. (2005).
\newblock Approximating conditional distribution functions using dimension
  reduction.
\newblock {\em The Annals of Statistics}, 33:1404--1421.

\bibitem[Hothorn and Zeileis, 2017]{hothorn2017transformation}
Hothorn, T. and Zeileis, A. (2017).
\newblock Transformation forests.
\newblock {\em arXiv preprint arXiv:1701.02110}.

\bibitem[Imbens and Rubin, 2015]{IR15}
Imbens, G.~W. and Rubin, D.~B. (2015).
\newblock {\em Causal Inference for Statistics, Social, and Biomedical
  Sciences: An Introduction}.
\newblock Cambridge University Press, Cambridge.

\bibitem[K{\"u}nzel et~al., 2019]{KSBY17}
K{\"u}nzel, S.~R., Sekhon, J.~S., Bickel, P.~J., and Yu, B. (2019).
\newblock Metalearners for estimating heterogeneous treatment effects using
  machine learning.
\newblock {\em Proceedings of the National Academy of Sciences},
  116:4156--4165.

\bibitem[Lakshminarayanan et~al., 2014]{MF}
Lakshminarayanan, B., Roy, D.~M., and Teh, Y.~W. (2014).
\newblock Mondrian forests: Efficient online random forests.
\newblock In Ghahramani, Z., Welling, M., Cortes, C., Lawrence, N.~D., and
  Weinberger, K.~Q., editors, {\em Advances in Neural Information Processing
  Systems 27}, pages 3140--3148. Curran Associates, Inc.

\bibitem[{Lin} et~al., 2019]{greenkhorn}
{Lin}, T., {Ho}, N., and {Jordan}, M.~I. (2019).
\newblock {On the efficiency of the Sinkhorn and Greenkhorn algorithms and
  their acceleration for optimal transport}.
\newblock {\em arXiv}, 1906.01437.

\bibitem[Meinshausen, 2006]{QRF}
Meinshausen, N. (2006).
\newblock Quantile regression forests.
\newblock {\em Journal of Machine Learning Research}, 7:983--999.

\bibitem[Miller et~al., 2014]{MRF2}
Miller, K., Huettmann, F., Norcross, B., and Lorenz, M. (2014).
\newblock Multivariate random forest models of estuarine-associated fish and
  invertebrate communities.
\newblock {\em Marine Ecology Progress Series}, 500:159--174.

\bibitem[{Nie} and {Wager}, 2017]{NW17}
{Nie}, X. and {Wager}, S. (2017).
\newblock {Quasi-oracle estimation of heterogeneous treatment effects}.
\newblock {\em arXiv}, 1712.04912.

\bibitem[{Peyr{\'e}} and {Cuturi}, 2018]{Wp-book}
{Peyr{\'e}}, G. and {Cuturi}, M. (2018).
\newblock {Computational Optimal Transport}.
\newblock {\em arXiv}, 1803.00567.

\bibitem[Pospisil and Lee, 2019]{RFCDE}
Pospisil, T. and Lee, A.~B. (2019).
\newblock {R}{F}{C}{D}{E}: Random {F}orests for {C}onditional {D}ensity
  {E}stimation and functional data.
\newblock {\em arXiv}, 1906.07177.

\bibitem[Rosenblatt, 1969]{rosenblatt1969conditional}
Rosenblatt, M. (1969).
\newblock Conditional probability density and regression estimators.
\newblock {\em Multivariate analysis II}, 25:31.

\bibitem[Rubin, 1974]{Rub74}
Rubin, D. (1974).
\newblock Estimating causal effects of treatments in randomized and
  nonrandomized studies.
\newblock {\em Journal of Educational Psychology}, 66:688--701.

\bibitem[Santambrogio, 2015]{santambrogio2015optimal}
Santambrogio, F. (2015).
\newblock {\em Optimal transport for applied mathematicians}, volume~55.
\newblock Springer.

\bibitem[Scornet et~al., 2015]{consistency-RF}
Scornet, E., Biau, G., and Vert, J.-P. (2015).
\newblock Consistency of random forests.
\newblock {\em The Annals of Statistics}, 43:1716--1741.

\bibitem[Segal and Xiao, 2011]{MRF1}
Segal, M. and Xiao, Y. (2011).
\newblock Multivariate random forests.
\newblock {\em WIREs Data Mining and Knowledge Discovery}, 1:80--87.

\bibitem[{Wager}, 2014]{Wager15}
{Wager}, S. (2014).
\newblock {Asymptotic theory for random forests}.
\newblock {\em arXiv}, 1405.0352.

\bibitem[Waskom et~al., 2020]{seaborn}
Waskom, M., Botvinnik, O., Ostblom, J., Gelbart, M., Lukauskas, S., Hobson, P.,
  Gemperline, D.~C., Augspurger, T., Halchenko, Y., Cole, J.~B., Warmenhoven,
  J., de~Ruiter, J., Pye, C., Hoyer, S., Vanderplas, J., Villalba, S., Kunter,
  G., Quintero, E., Bachant, P., Martin, M., Meyer, K., Swain, C., Miles, A.,
  Brunner, T., O'Kane, D., Yarkoni, T., Williams, M.~L., Evans, C., Fitzgerald,
  C., and Brian (2020).
\newblock mwaskom/seaborn: v0.10.1 (april 2020).

\end{thebibliography}
\end{document}